\documentclass[]{article}

\usepackage[margin=1in]{geometry}  
\usepackage{multirow} 
\usepackage{graphicx}              
\usepackage{epstopdf}
\usepackage{amsmath}               
\usepackage{amsfonts}              
\usepackage{amsthm}                
\usepackage{subfigure}                
\usepackage{url}
\usepackage{graphicx}
\usepackage{hyperref}
\begin{document}

\date{}


\title{Feasibility Preserving Constraint-Handling Strategies for Real Parameter Evolutionary Optimization}

\author{\bf{Nikhil Padhye} \\
{npdhye@mit.edu}\\ 
Department of Mechanical Engineering,\\
Massachusetts Institute of Technology\\ 
Cambridge, MA-02139, USA\\
\bf{Pulkit Mittal}\\
{pulkitm.iitk@gmail.com}\\
Department of Electrical Engineering\\
Indian Institute of Technology Kanpur\\ 
Kanpur-208016, U.P., India\\
\bf{Kalyanmoy Deb}\\
kdeb@egr.msu.edu	\\
Department of Electrical and Computer Engineering\\
Michigan State University \\
East Lansing, MI 48824, USA\\
}
 
\maketitle

\begin{abstract}
Evolutionary Algorithms (EAs) are being routinely applied for 
a variety of optimization tasks, and real-parameter optimization
in the presence of constraints is one such important area. During constrained optimization EAs
often create solutions that fall outside the feasible region; hence a viable constraint-handling strategy is needed. 
This paper focuses on the class of constraint-handling strategies that repair 
infeasible 
solutions by bringing them back into the search space and explicitly preserve feasibility 
of the solutions. Several existing constraint-handling
strategies are studied, and two new single parameter constraint-handling methodologies based on 
parent-centric and inverse parabolic probability (IP) distribution are proposed. 
The existing and
newly proposed constraint-handling methods are first studied with PSO, DE, GAs,
and simulation results on four scalable test-problems 
under different location settings of the optimum are presented. The newly proposed constraint-handling methods
exhibit robustness in terms of performance and also succeed on search spaces 
comprising up-to $500$ variables while locating the optimum within an error of 
$10^{-10}$. The working principle of the IP based methods is 
also demonstrated on (i) some generic constrained optimization problems,
and (ii) a classic `Weld' problem from structural design and mechanics. The 
successful performance of the proposed methods 
clearly exhibits their efficacy as a generic constrained-handling strategy for 
a wide range of applications.  
\end{abstract}

\textbf{Keywords}
constraint-handling,
nonlinear and constrained optimization,
particle swarm optimization,
real-parameter genetic algorithms,
differential evolution.

\section{Introduction}
Optimization problems are wide-spread in several domains of science and engineering. 
The usual goal is to minimize or maximize some pre-defined objective(s). 
Most of the real-world scenarios place certain restrictions 
on the variables of the problem i.e. the variables need to satisfy 
certain pre-defined constraints to realize an acceptable solution.
 
The most general form of a constrained optimization problem (with inequality constraints,
equality constraints and variable bounds) can be written as a nonlinear programming
(NLP) problem: 

{\small\begin{eqnarray}
Minimize	 &f(\vec{x})		&	\nonumber\\
Subject to	 &g_j(\vec{x}) \ge 0,   & j= 1,...,J \nonumber\\
		 &h_k(\vec{x})  =  0,	& k= 1,...,K \nonumber\\
	  	 & x_i^{(L)} \le x_i \le x_i^{(U)}, & i=1,...,n. \label{eq:NLPprob}
\end{eqnarray}}

The NLP problem defined above contains $n$ decision variables (i.e. $\vec{x}$ is a vector of size $n$),
$J$ greater-than-equal-to type inequality constraints (less-than-equal-to can be expressed in this form by 
multiplying both sides by $-1$), and $K$ equality-type constraints.
The problem variables $x_is$ are bounded by the lower ($x_i^{(L)}$) and upper ($x_i^{(U)}$)
limits. When only the variable bounds are specified then the 
constraint-handling strategies are often termed as the boundary-handling methods.
\footnote{For the rest of the paper, by \textit{constraint-handling} we imply
tackling all of the following: variable bounds, inequality constraints and equality constraints. 
And, by a \textit{feasible solution} it is implied that the solution satisfies all the variable bounds, inequality constraints,
and equality constraints. The main contribution of the paper is to propose an efficient constraint-handling method
that operates and generates only feasible solutions during optimization.}

In classical optimization, the task of constraint-handling has been addressed
in a variety of ways: (i) \textit{using penalty approach} developed by Fiacoo and McCormick \cite{jensen2003operations},
which degrades the function value
in the regions outside the feasible domain, (ii) \textit{using barrier methods} which operate in a similar fashion 
but strongly degrade the function values as the solution approaches a constraint boundary from 
inside the feasible space, (iii) \textit{performing search in the feasible directions} using methods
such gradient projection, reduced gradient and Zoutendijk's approach \cite{zoutendyk1960methods}
(iv) \textit{using the augmented Lagrangian formulation} of the problem, as commonly 
done in linear programming and sequential quadratic programming (SQP).
For a detailed account on these methods along with their implementation and 
convergence characteristics the reader is referred to \cite{rekl,debOptiBOOK,MichalewiczConstraint}.
The classical optimization methods reliably and effectively solve convex constrained optimization problems while ensuring convergence
and therefore widely used in such scenarios.  However, same is not true in the presence of non-convexity.
The goal of this paper is to address the issue of constraint-handling for evolutionary algorithms in real-parameter optimization, 
without any limitations to convexity or a special form of constraints or objective functions. 

In context to the evolutionary algorithms the constraint-handling has been addressed
by a variety of methods; including borrowing of the ideas from the classical techniques. These include
(i) \textit{use of penalty functions} to degrade the fitness
values of infeasible solutions such that the degraded solutions are given less emphasis during the evolutionary search. A
common challenge in employing such penalty methods arises from
choosing an appropriate penalty parameter ($R$) that strikes the right balance between
the objective function value, the amount of constraint violation and the associated penalty. 
Usually, in EA studies, a trial-and-error method is employed to estimate $R$.
A study \cite{debpenalty} in 2000 suggested a parameter-less
approach of implementing the penalty function concepts for population-based optimization method. 
A recent bi-objective method
\cite{deb-dutta} was reported to find the appropriate $R$ values adaptively during the optimization process.
Other studies \cite{wang2012dynamic,wang2012combining} have employed the concepts of multi-objective
optimization by simultaneously considering the minimization of the constraint violation and optimization of the
objective function, (ii) \textit{use of feasibility preserving operators},
for example, in \cite{michalewicz1996genocop} specialized operators in the presence of linear constraints 
were proposed to create new and feasible-only individuals from the feasible parents. In another example, 
generation of feasible child solutions within the variable bounds was achieved  
through Simulated Binary Crossover (SBX) \cite{debSBX} and polynomial mutation
operators \cite{debBookMO}. The explicit feasibility of child solutions was ensured by redistributing the probability distribution 
function in such a manner that the infeasible regions were assigned a zero probability 
for child-creation \cite{debpenalty}. Although explicit creation of feasible-only solutions during an EA search is an
attractive proposition, but it may not be possible always since generic crossover or mutation operators
or other standard EAs do not gaurantee creation of feasible-only solutions, (iii) \textit{deployment of repair strategies}
that bring an infeasible solution back into the feasible domain.
Recent studies \cite{PadhyeBIC2012,Helwid-Constraing-handling-2013,amir,chu} investigated the 
issue of constraint-handling through repair techniques in context to PSO and DE, and   
showed that the repair mechanisms can introduce a bias in the search and 
hinder exploration. Several repair methods 
proposed in context PSO \cite{padhyeCEC2009,Sabina,JonathanPareto} 
exploit the information about location of the optimum and fail to perform when the location of 
optimum changes \cite{padhye2010}. These issues are universal and
often encountered with all EAs (as shown in later sections).
Furthermore, the choice of the evolutionary optimizer, the constraint-handling strategy, and 
the location of the optima with respect to the search space, all play an important
role in the optimization task. To this end, authors have realized a need
for a reliable and effective repair-strategy that explicitly preserves feasibility.
An ideal evolutionary optimizer (evolutionary algorithm
and its constrained-handling strategy) should be robust in terms of finding the 
optimum, irrespective of the location of the optimal location in the search space.   
In rest of the paper, the term constraint-handling strategy refers to explicit feasibility
preserving repair techniques.  

First we review the existing constraint-handling strategies 
and then propose two new constraint-handling schemes, namely, Inverse Parabolic Methods (IPMs). 
Several existing and newly proposed constrained-handling strategies are first tested on a class 
of benchmark unimodal problems with variable bound constraints.
Studying the performance of constraint-handling strategies on problems with variable bounds
allows us to gain better understanding into the operating principles in a simplistic manner. 
Particle Swarm Optimization, Differential Evolution and real-coded Genetic Algorithms are chosen 
as evolutionary optimizers to study the performance of different constraint-handling strategies. 
By choosing different evolutionary optimizers, better understanding on the functioning
of constraint-handlers embedded in the evolutionary frame-work can be gained. 
Both, the search algorithm and constraint-handling strategy must operate efficiently and synergistically in
order to successfully carry out the optimization task. It is shown that the constraint-handling
methods possessing inherent pre-disposition; in terms of bringing infeasible solutions back into the
specific regions of the feasible domain, perform poorly. Deterministic constraint-handling strategies such as 
those setting the solutions on the constraint boundaries result in the loss of population diversity. 
On the other hand, random methods of bringing the solutions back into the search space
arbitrarily; lead to complete loss of all useful information carried by the solutions. 
A balanced approach that utilizes the useful information from the solutions
and brings them back into the search space in a meaningful way is desired. The newly proposed IPMs are motivated
by these considerations.
The stochastic and adaptive components of IPMs (utilizing the information of the solution's
feasible and infeasible locations), and a user-defined parameter ($\alpha$) render
them quite effective. 
  
The rest of the paper is organized as follows:
Section~\ref{sec:feasibility-preserving-existing} reviews existing constraint-handling techniques
commonly employed for problems with variable bounds.
Section~\ref{sec:IP} provides a detailed description on two newly proposed IPMs.
Section~\ref{sec:ResultsDiscussion} provides a description on the benchmark test problems and 
several simulations performed on PSO, GAs and DE with different constraint-handling techniques. 
Section~\ref{sec:scale-up} considers 
optimization problems with larger number of variables. 
Section~\ref{sec:Constraint-Programming} shows the extension and applicability of proposed
IPMs for generic constrained problems.
Finally, conclusions and scope for future work are discussed in Section~\ref{sec:Conclusion}.
    


\section{Feasibility Preserving Constraint-Handling Approaches for Optimization Problems with Variable Bounds}
\label{sec:feasibility-preserving-existing}
Several constraint-handling strategies have been proposed to bring solutions back into the feasible region
when constraints manifest as variable bounds. Some of these strategies can also be extended in 
presence of general constraints. An exhaustive recollection and 
comparison of all the constraint-handling techniques is beyond the scope of this study. 
Rather, we focus our discussions on the popular and representative constraint-handling techniques. 

The existing constraint-handling methods for problems with variable bounds can be broadly categorized into two groups: 
Group~$A$ techniques that perform feasibility check variable wise, and Group~$B$ techniques that perform feasibility
check vector-wise. According to Group~$A$ techniques, for every solution, each variable is tested for its feasibility 
with respect to its supplied bounds and made feasible if the corresponding bound is violated.
Here, only the variables violating their corresponding bounds are altered, independently, and
other variables are kept unchanged.  
According to Group~$B$ techniques, if a solution (represented as a vector) 
is found to violate any of the variable bounds, it is brought back into
the search space along a vector direction into the feasible space. In
such cases, the variables that explicitly do not violate their own
bounds may also get modified.
 
It is speculated that for variable-wise separable problems, that is,
problems where variables are not linked to one another, 
techniques belonging to Group~$A$ are likely to perform well. However, for the problems  
with high correlation amongst the variables (usually referred to as {\em linked}-problems), Group~$B$ techniques are likely to be more useful. 
Next, we provide description of these constraint-handling methods in detail \footnote{The implementation of several 
strategies as C codes can be obtained by emailing npdhye@gmail.com or pulkitm.iitk@gmail.com}.

\subsection{Random Approach}

\begin{figure}[hbt]
\begin{center}
\includegraphics[scale=.35]{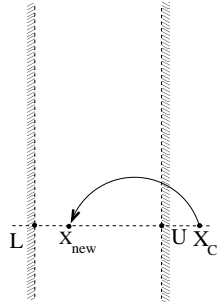}    
\end{center}
\caption{Variable-wise random approach for handling bounds.}
\label{fig:RandomBH}
\end{figure}
This is one of the simplest and commonly used approaches for handling boundary
violations in EAs \cite{chu}. This approach belongs to Group~$A$. 
Each variable is checked for a boundary violation and if the variable bound is violated
by the current position, say $x_i^c$, then $x_i^c$ is replaced with a randomly chosen value
$y_i$ in the range $[x_i^{(L)}, x_i^{(U)}]$, as follows:
\begin{equation}
y_i = \mbox{random} [x_i^{(L)}, x_i^{(U)}].
\end{equation}
Figure~\ref{fig:RandomBH} illustrates this
approach. Due to the random choice of the feasible location, this approach explicitly maintains
diversity in the EA population. 

\subsection{Periodic Approach}
\begin{figure}[hbt]
\begin{center}
\includegraphics[scale=.35]{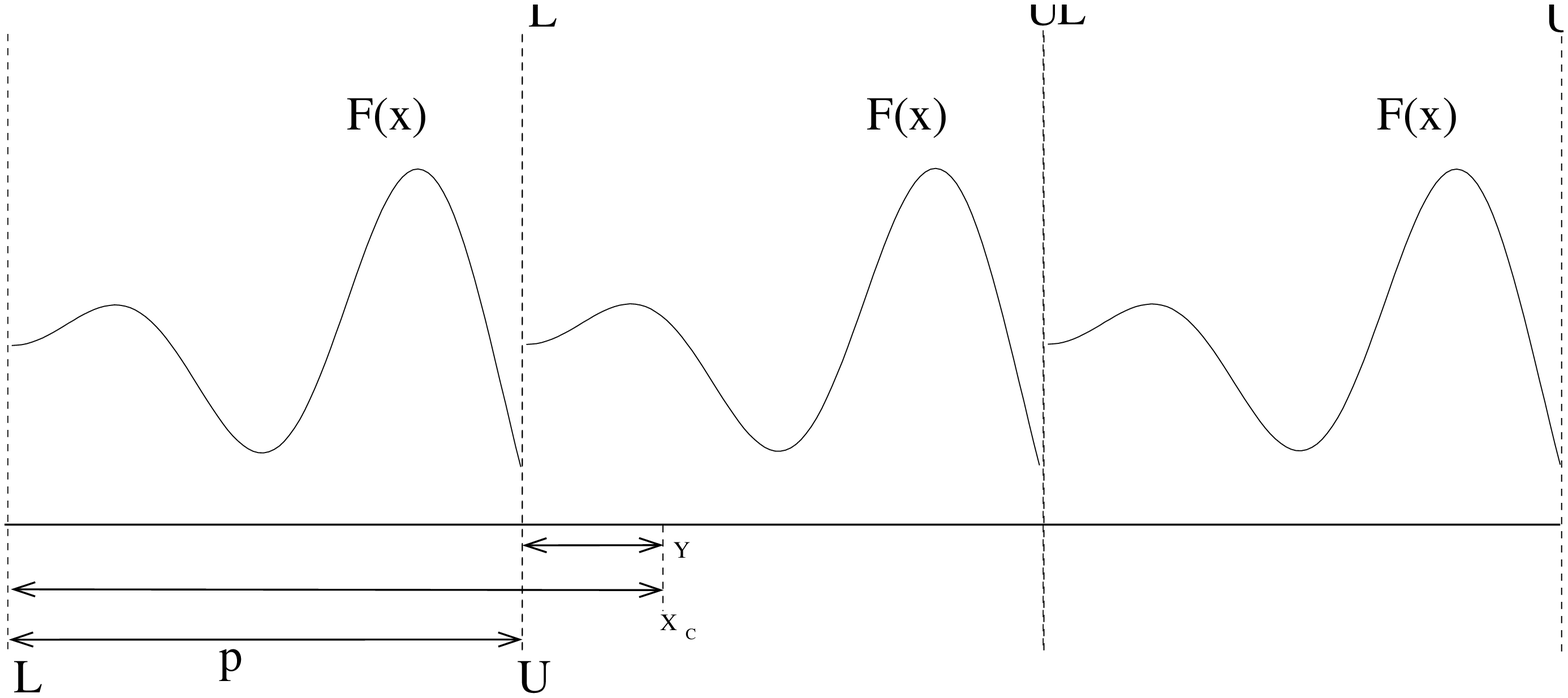}    
\end{center}
\caption{Variable-wise periodic approach for handling bounds.}
\label{fig:PeriodicBH}
\end{figure}
This strategy assumes a periodic repetition of the objective function
and constraints with a period $p=x_i^{(U)}-x_i^{(L)}$. This is carried out by
mapping a violated variable $x_i^c$ in the range $[x_i^{(L)},
x_i^{(U)}]$ to $y_i$, as follows:
\begin{equation}
y_i  = \left\{ 
\begin{array}{ll}
         x_i^{(U)} - (x_i^{(L)}-x_i^c)\%p, & \quad \text{if $x_i^c<x_i^{(L)}$},\\
         x_i^{(L)} + (x_i^c-x_i^{(U)})\%p, & \quad \text{if $x_i^c>x_i^{(U)}$}, \\
\end{array} 
\right.
\end{equation}
In the above equation, \% refers to the modulo operator.
Figure~\ref{fig:PeriodicBH} describes the periodic approach. 
The above operation brings back an infeasible solution in a structured
manner to the feasible region. 
In contrast to the random method, the periodic approach is too methodical and it is unclear 
whether such a repair mechanism is supportive of
preserving any meaningful information of the solutions that have created the
infeasible solution. This approach belongs to Group~$A$. 

\subsection{SetOnBoundary Approach}
As the name suggests, according to this strategy a violated variable
is reset on the 
bound of the variable which it violates.
\begin{equation}
y_i  = \left\{ 
\begin{array}{ll}
         x_i^{(L)}, & \quad \text{if $x_i^c<x_i^{(L)}$},\\
         x_i^{(U)}, & \quad \text{if $x_i^c>x_i^{(U)}$}.\\
\end{array} \right.
\end{equation}
Clearly this approach forces all violated solutions to lie on the
lower or on the upper boundaries, as the case may be. Intuitively, this approach will work
well on the problems when the optimum of the problem lies exactly on one of the variable
boundaries. This approach belongs to Group~$A$.

\subsection{Exponentially Confined (Exp-C) Approach}
\begin{figure}[hbt]
\begin{minipage}{0.47\linewidth}
\begin{center}
\includegraphics[scale=.35]{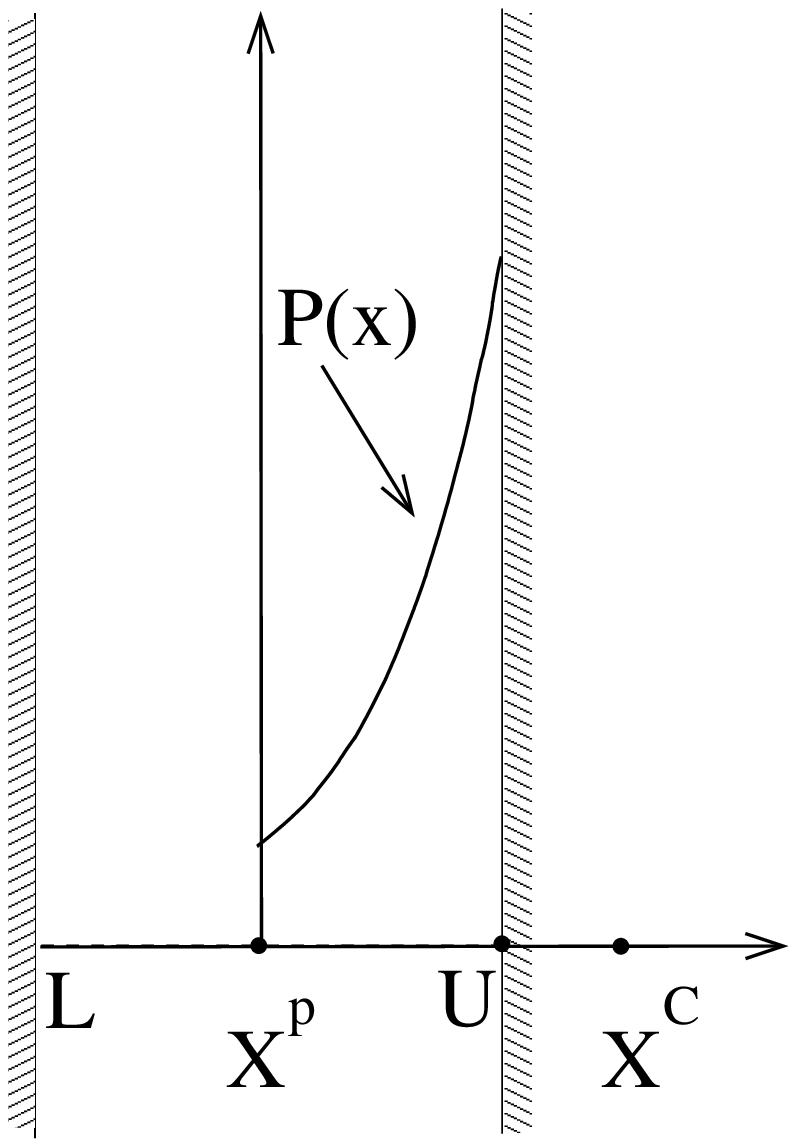}    
\end{center}
\caption{Variable-wise exponentially approach (Exp-C) for handling
  bounds.}
\label{fig:FieldSendDistBH}
\end{minipage}\hfill
\begin{minipage}{0.47\linewidth}
\begin{center}
\includegraphics[scale=.35]{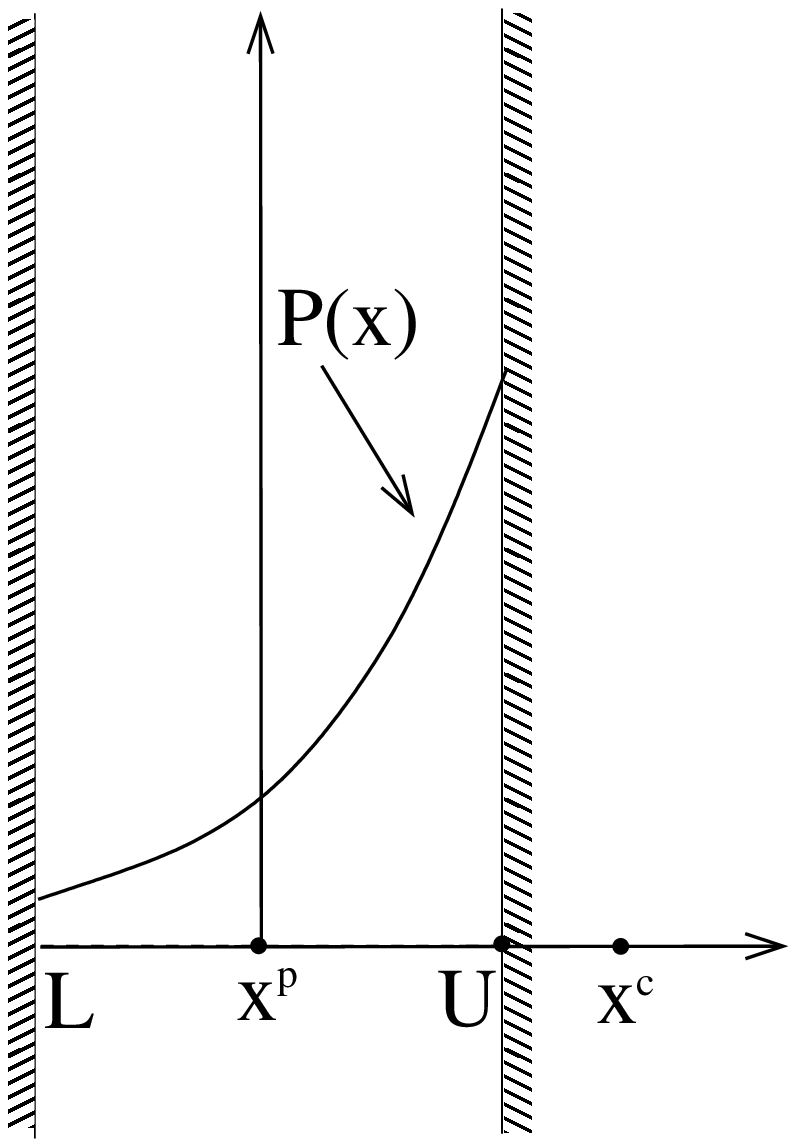}    
\end{center}
\caption{Variable-wise exponentially approach (Exp-S) for handling bounds.}
\label{fig:expS}
\end{minipage}
\end{figure}
This method was proposed in \cite{JonathanPareto}. According to this 
approach, a particle is brought back inside the feasible search space variable-wise in the region between 
its old position and the violated bound. The new location is created
in such a manner that higher sampling probabilities are assigned to the regions
near the violated boundary. The developers suggested the use of an
exponential probability distribution, shown in
Figure~\ref{fig:FieldSendDistBH}. 
The motivation of this approach is based on the hypothesis 
that a newly created infeasible point violates a particular variable
boundary because the optimum solution lies closer to that variable
boundary. Thus, this method will probabilistically 
create more solutions closer to the boundaries, unless the optimum lies well
inside the restricted search space. This approach belongs to Group~$A$.

Assuming that the exponential distribution is $p(x_i) =
A\exp(|x_i-x_i^p|)$, the value of $A$ can be obtained by integrating
the probability from $x_i=x_i^p$ to $x_i=x_i^{(B)}$ (where $B=L$ or
$U$, as the case may be). Thus, the probability distribution is given
as $p(x) = \exp(|x_i-x_i^p|)/(\exp(|x_i^{(B)} - x_i^p|)-1)$. For any
random number $r$ within $[0,1]$, the feasible solution is calculated as follows:
\begin{equation}
y_i= \left\{\begin{array}{ll}
x_i^p - \ln (1+r(\exp (x_i^p-x_i^{(L)})-1))  &\mbox{if $x_i<x_i^{(L)}$},\\
x_i^p + \ln (1+r(\exp (x_i^{(U)}-x_i^p)-1)), & \mbox{if
  $x_i>x_i^{(U)}$}.
\end{array}\right.
\label{eq:exp}
\end{equation}

\subsection{Exponential Spread (Exp-S) Approach}
This is a variation of the above approach, in which, instead of 
confining the probability to lie between $x_i^p$ and the violated
boundary, the exponential probability is spread over the entire
feasible region, that is, the probability is distributed from lower
boundary to the upper boundary with an increasing probability towards
the violated boundary. This requires replacing $x_i^p$ with
$x_i^{(U)}$ (when the lower boundary is violated) or $x_i^{(L)}$ 
(when the upper boundary is violated) in the Equation~\ref{eq:exp} as follows: 
\begin{equation}
y_i= \left\{\begin{array}{ll}
x_i^{(U)} - \ln (1+r(\exp(x_i^{(U)}-x_i^{(L)})-1))  &\mbox{if $x_i<x_i^{(L)}$},\\
x_i^{(L)} + \ln (1+r(\exp(x_i^{(U)}-x_i^{(L)})-1)), & \mbox{if
  $x_i>x_i^{(U)}$}.
\end{array}\right.
\end{equation}
The probability distribution is shown in Figure~\ref{fig:expS}.
This approach also belongs to Group~$A$.

\subsection{Shrink Approach}
\begin{figure}[hbt]
\begin{center}
\includegraphics[scale=.35]{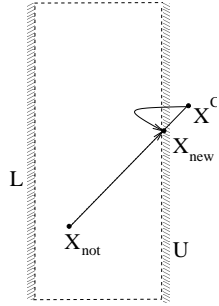}    
\end{center}
\caption{Vector based {SHR.} strategy for handling bounds.}
\label{fig:SHRBH}
\end{figure}
This is a vector-wise approach and belongs to Group~$B$ in which the violated solution is
set on the intersection point of the line joining the parent point
($\vec{x}_{not}$), child point ($\vec{x}^c)$, and the violated boundary. Mathematically,
the mapped vector $\vec{y}$ is created as follows:
\begin{equation}
\vec{y} = \vec{x}_{not} +\beta (\vec{x}^c-\vec{x}_{not}), 
\end{equation}
where $\beta$ is computed as the minimum of all positive values of intercept
$(x_i^{(L)}-x_{i,not})/(x_i^c-x_{i,not})$ for a violated boundary
$x_i^{(L)}$ and $(x_i^{(U)}-x_{i,not})/(x_i^c-x_{i,not})$ for a violated boundary
$x_i^{(U)}$.
This operation is shown in Figure~\ref{fig:SHRBH}. In the case shown,
$\beta$ needs to be
computed for variable bound $x_2^{(U)}$ only. \\ \\


\section{Proposed Inverse Parabolic (IP) Constraint-Handling Methods}\label{sec:IP}
\begin{figure}[hbt]
\begin{center}
\includegraphics[scale=.75]{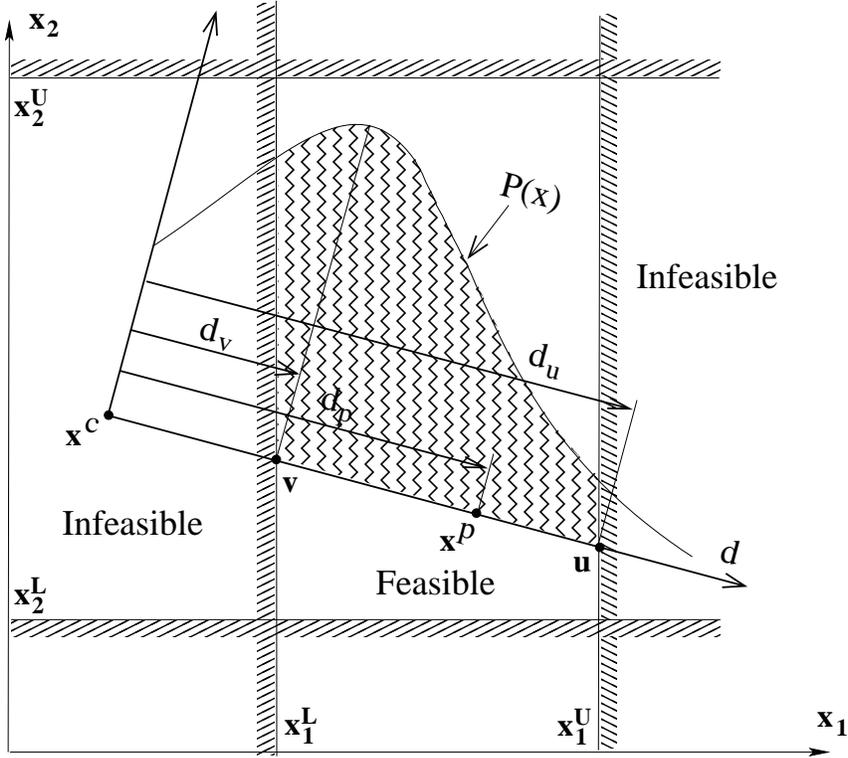}    
\end{center}
\caption{Vector based {Inverse Parabolic Methods}.}
\label{fig:ProbDistBH}
\end{figure}
The exponential probability distribution function described in the previous section brings 
violated solutions back into the allowed range variable-wise, but 
ignores the distance of the violated solution $x_i^c$ with respect to
the violated boundary. The distance from the violated boundary 
carries useful information for remapping the violated solution into
the feasible region. One way to utilize this distance information is
to bring solutions back into the allowed range with a higher
probability closer to the boundary, when the \textit{fallen-out}
distance ($d_v$, as shown in Figure~\ref{fig:ProbDistBH}) is small. 
In situations, when points are too far outside the allowable range,
that is, the fallen-out
distance $d_v$ is large, particles are brought back more uniformly
inside the feasible range. Importantly, when the fallen-out distance
$d_v$ is small (meaning that the violated child solution is close to the
variable boundary), the repaired point is also close to the violated
boundary but in the feasible side. Therefore, the nature of the exponential
distribution should become more and more like a uniform distribution
as the fallen-out distance $d_v$ becomes large.

Let us consider Figure~\ref{fig:ProbDistBH} which shows a
violated solution $\vec{x}^c$ and its parent solution $\vec{x}^p$. Let
$d_p=\|\vec{x}^c-\vec{x}^p\|$ denote the distance between the violated solution
and the parent solution. Let $\vec{v}$ and $\vec{u}$ be the intersection points
of the line joining $\vec{x}^c$ and $\vec{x}^p$ with the violated
boundary and the non-violated boundary, respectively. The
corresponding distances of these two points from $\vec{x}^c$ are $d_v$
and $d_u$, respectively. Clearly, the violated distance is $d_v =
\|\vec{x}^c-\vec{v}\|$.  We now define an inverse
parabolic probability distribution function from $\vec{x}^c$ along 
the direction $(\vec{x}^p-\vec{x}^c)$ as:
\begin{equation}
p(d) = \frac{A}{(d-d_v)^2+\alpha^2d_v^2}, \quad d_v \leq d \leq a,
\end{equation}
where $a$ is the upper bound of $d$ allowed by the constraint-handling
scheme (we define $a$ later) and $\alpha$ is a pre-defined parameter. By calculating and
equating the cumulative probability equal to one, we find:
\[A = \frac{\alpha d_v}{\tan^{-1} \frac{a-d_v}{\alpha d_v}}.\]
The probability is maximum at $d=d_v$ (at the violated boundary) and
reduces as the solution enters the allowable range. Although this 
characteristic was also present in the exponential distribution, the
above probability distribution is also a function of violated distance
$d_v$, which acts
like a variance to the probability distribution. If $d_v$ is
small, then the variance of the distribution is small, thereby resulting in a
localized effect of creating a mapped solution. 
For a random number $r\in [0,1]$, the distance of the mapped solution
from $\vec{x}^c$ in the allowable
range $[d_v,d_u]$ is given as follows:
\begin{equation}
d' = d_v + \alpha d_v \tan \left(r \tan^{-1} \frac{a-d_v}{\alpha d_v}\right).
\label{eq:s}
\end{equation}
The corresponding mapped solution is as follows:
\begin{equation}
\vec{y} = \vec{x}^c + d' (\vec{x}^p-\vec{x}^c).
\label{eq:map}
\end{equation}
Note that the IP method makes a vector-wise operation and is sensitive
to the relative locations of the infeasible solution, the parent solution, and
the violated boundary. 

The parameter $\alpha$ has a direct {\em external\/} effect of inducing small or large
variance to the above probability distribution. If $\alpha$ is large,
the variance is large, thereby having uniform-like distribution. Later 
we shall study the effect of the parameter $\alpha$. A value $\alpha$ $\approx$ 1.2 is
found to work well in most of the problems and is recommended. 
Next, we describe two particular constraint-handling schemes employing this  
probability distribution. 

\subsection{Inverse Parabolic Confined (IP-C) Method}
In this approach, the probability distribution is confined
between $d \in [d_v,d_p]$, thereby making $a=d_p$. Here, a mapped
solution $\vec{y}$ lies strictly between violated boundary location
($\vec{v}$) and the parent ($\vec{x}^p$). 

\subsection{Inverse Parabolic Spread (IP-S) Method} 
Here, the mapped solution is allowed to lie in the entire
feasible range  between $\vec{v}$ and
$\vec{u}$ along the vector $(\vec{x}^p-\vec{x}^c)$, but more emphasis is given on relocating the 
child near the violated boundary. The solution can be found by using Equations~\ref{eq:s} and
\ref{eq:map}, and by setting $a=d_u$. 

\section{Results and Discussions}\label{sec:ResultsDiscussion}
In this study, first we choose four standard scalable unimodal test functions (in presence of variable bounds): 
Ellipsoidal ($F_{\rm elp}$), Schwefel ($F_{\rm sch}$), Ackley ($F_{\rm sch}$),
and Rosenbrock ($F_{\rm ros}$), described as follows:
\begin{eqnarray}
F_{\rm elp} &=& \sum_{i=1}^n ix_i^2 \\
F_{\rm sch} &=& \sum_{i=1}^n\left(\sum_{j=1}^i x_j\right)^2 \\
F_{\rm ack} &=& -20 exp \left(-0.2 \sqrt{ \frac{1}{n} \sum_{i=1}^{i=n} x_i^2}\right) -exp\left(\frac{1}{n} \sum_{i=1}^{n}cos(2\pi x_i)\right)  +20 + e\\
F_{\rm ros} &=& \sum_{i=1}^{n-1} (100(x_i^2-x_{i+1})^2  + (x_i-1)^2) 
\end{eqnarray}

In the unconstrained space, $F_{\rm elp}$, $F_{\rm sch}$ and $F_{\rm ack}$ have a minimum
at $x_i^{\ast}=0$, whereas $F_{ros}$ has a minimum at $x_i^{\ast}=1$.
All functions have minimum value $F^{\ast}=0$.  
$F_{\rm elp}$ is the only variable separable problem.
$F_{\rm ros}$ is a challenging test problem that has a ridge which poses
difficulty for several optimizers. 
In all the cases the number of variables is chosen to be $n=20$. 

\begin{figure}
\centering
\begin{subfigure}{}
  \centering
  \includegraphics[width=.65\linewidth]{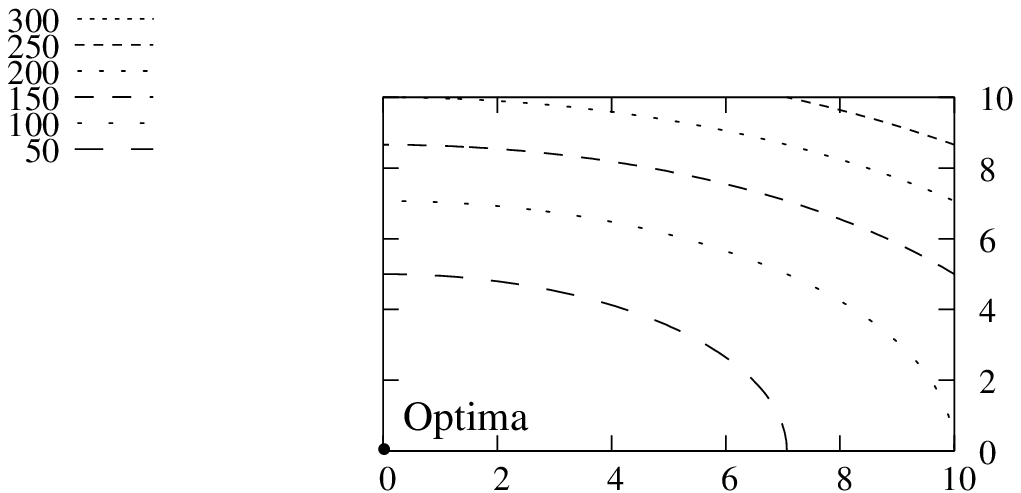}
  \label{fig:boundary-ellp-optima}
\end{subfigure}%

\begin{subfigure}{}
  \centering
  \includegraphics[width=.65\linewidth]{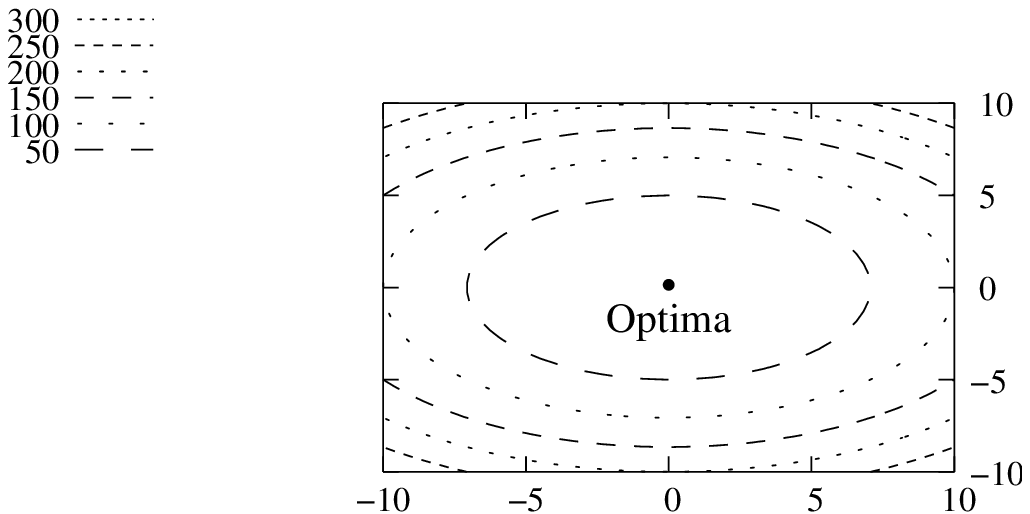}
  \label{fig:center-ellp-optima}
\end{subfigure}

\begin{subfigure}{}
  \centering
  \includegraphics[width=.65\linewidth]{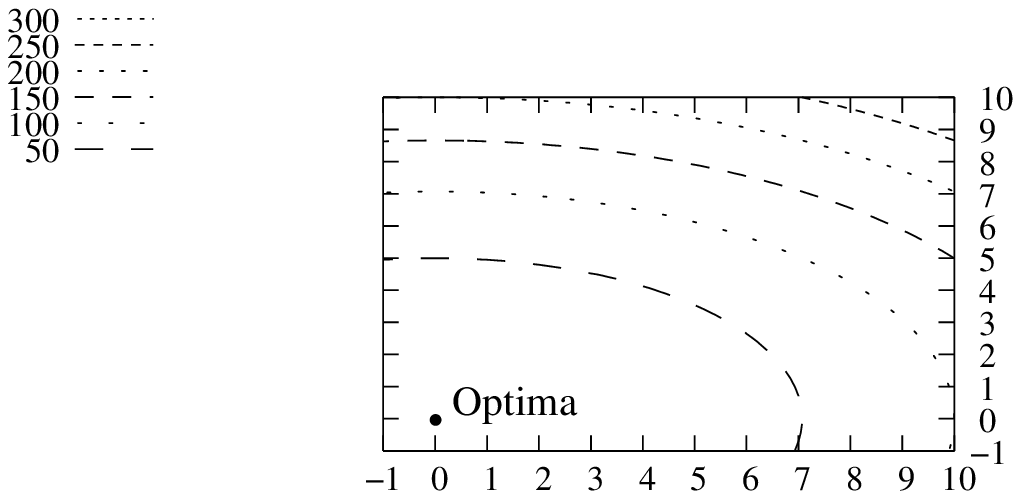}
  \label{fig:close-to-boundary-ellp-optim}
\end{subfigure}

\caption{Location of optimum for $F_{elp}$: (a) on the boundary (b) in the center and (c) close to the 
edge of the boundary by selecting different search domains.}
\label{fig:3scenario}
\end{figure}

For each test problem three different scenarios corresponding to the
relative location of the 
optimum with respect to the allowable search range are considered. This is done 
by selecting different variable bounds, as follows:  
 
\begin{description}
\item[On the Boundary:] Optimum is exactly on one of the variable boundaries (for $F_{\rm elp}$, $F_{\rm sch}$ and $F_{\rm ack}$ 
$x_i\in [0, 10]$, and for $F_{\rm ros}$, $x_i\in [1, 10]$),
\item[At the Center:] Optimum is at the center of the allowable range (for $F_{\rm elp}$, $F_{\rm sch}$ and $F_{\rm ack}$ 
$x_i\in [-10, 10]$, and for $F_{\rm ros}$, $x_i\in [-8, 10]$), and 
\item[Close to Boundary:] Optimum is near the variable boundary, 
but not exactly on the boundary (for $F_{\rm elp}$, $F_{\rm sch}$ and $F_{\rm ack}$ 
$x_i\in [-1, 10]$, and for $F_{\rm ros}$,  $x_i\in [0, 10]$).
\end{description}
These three scenarios are shown in the Figure~\ref{fig:3scenario} for a
two-variable problem having variable bounds: $x_i^{(L)}=0$ and $x_i^{(U)}=10$.
Although in practice, the optimum can lie anywhere in the allowable range,
the above three scenarios pose adequate representation of different
possibilities that may exist in practice.  

For each test problem, the population is initialized uniformly in the
allowable range. We count the number of function
evaluations needed for the algorithm to find a solution close to the
known optimum solution and we call this our 
evaluation criterion $S$.  
Choosing a high accuracy (i.e. small value of $S$)
as the termination criteria minimizes the chances of locating the optimum due to 
random effects, and provides a better insight into the behavior of a constraint-handling mechanism.  

To eliminate the random effects and gather results of statistical importance, each algorithm 
is tested on a problem $50$ times (each run starting with a different
initial population). A particular run is terminated if the evaluation
criterion $S$ is met (noted as a successful run), 
or the number of function evaluations exceeds one million (noted as an
unsuccessful run). If only a few out of the $50$ runs are
successful, then we report the number of successful runs in the
bracket. In this case, the best, median and worst number of function
evaluations are computed from the successful runs only. If none of the runs
are successful, we denote this by marking \textit{(DNC)} (Did Not
Converge). In such cases, we report the best, median and worst
attained function values of the best solution at the end of each
run.  To distinguish the unsuccessful results from successful
ones, we present the fitness value information of the unsuccessful
runs in italics.

An in-depth study on the constraint-handling techniques is carried out
in this paper. 
Different locations of the optimum are selected and systematic comparisons are 
carried out for PSO, DE and GAs in Sections ~\ref{subsec:psoresults}, 
~\ref{subsec:DEresults} and ~\ref{subsec:GAresults} , respectively. 
 
\subsection{Results with Particle Swarm Optimization (PSO)}\label{subsec:psoresults}
In PSO, decision variable and the velocity terms are updated
independently. Let us say, that the initial position is $\vec{x}_{t}$, the newly created 
position is infeasible and represented by $\vec{x}_{t+1}$, and the repaired solution
is denoted by  $\vec{y}_{t}$. 

If the velocity update is based on the infeasible solution as:

\begin{equation}
\label{eqn:velocity-standard}
\vec{v}_{t+1}=\vec{x}_{t+1} - \vec{x}_{t} 
\end{equation}

then, we refer to this as ``Velocity Unchanged''. However, if the velocity update 
is based on the repaired location as: 

\begin{equation}
\label{eqn:velocity-recomputed}
\vec{v}_{t+1}=\vec{y}_{t} - \vec{x}_{t},  
\end{equation}

then, we refer to this as ``Velocity Recomputed''. This terminology is used for rest of the paper. 
For inverse parabolic (IP) and exponential (Exp) approaches, we use
``Velocity Recomputed'' strategy only. We have performed ``Velocity
Unchanged'' strategy with IP and exponential approaches, but the
results were not as good as compared to ``Velocity Recomputed'' strategy. 
For the \textit{SetOnBoundary} approach, we use the ``Velocity Recomputed''
strategy and two other strategies discussed as follows. 

Another strategy named 
``Velocity Reflection'' is used, which simply implies 
that if a particle is set on the $i$-th boundary, then $v_i^{t+1}$ 
is changed to $-v_i^{t+1}$. The goal of the velocity
reflection is  to explicitly allow particles to move back into the search space.
In the ``Velocity Set to Zero'' strategy,  if a particle is set
on the $i$-th boundary, then the corresponding velocity component is set to zero i.e. $v_i^{t+1}=0$. 
For the shrink approach, both ``Velocity Recomputed'' and ``Velocity
Set to Zero'' strategies are used. 

For PSO, a recently proposed {\em hyperbolic} \cite{Helwid-Constraing-handling-2013} constraint-handling approach is also
included in this study. This strategy operates by first calculating velocity according to the standard mechanism ~\ref{eqn:velocity-standard}, and 
in the case of violation a linear normalization is performed on the velocity to restrict the solution from jumping out of the constrained boundary as follows:
\begin{equation}
\label{eq:hyperbolic}
v_{i,t+1} =  \frac{v_{i,t+1}}{1+\frac{|v_{i,t+1}|}{\min(x_i^{(U)}-x_{i},x_{i,t}-x_{i}^{(L)})}}. 
\end{equation}
Essentially, the closer the particle gets to the boundary (e.g.,
$x_{i,t}$ only slightly smaller than $x_{i}^{(U)}$), the more difficult it becomes to reach the boundary. In fact, the particle is never completely
allowed to reach the boundary as the velocity tends to zero. We emphasize again that this strategy is only applicable to
PSO. A standard PSO is employed in this study with a population size of 100. 
The results for all the above scenarios with PSO are presented in
Tables~\ref{tab:PSOEllp} to ~\ref{tab:PSORos}.

\begin{table*}[ht]
\begin{footnotesize}
\caption{Results on $F_{\rm elp}$ with PSO for $10^{-10}$ termination criterion.}
\begin{minipage}[b]{1.0\linewidth}
\begin{center}
\begin{tabular}{|lrrrr|} \hline 
{Strategy}   & Velocity update             &  {Best} &
 Median &  Worst \\ \hline \hline 
\multicolumn{5}{|c|}{	{$F_{\rm elp}$ in [0,10]: On the Boundary}} \\ \hline
IP Spread   & Recomputed          & 39,900  &    47,000    &
67,000    \\ IP Confined & Recomputed            & 47,900 (49) &   88,600   &  140,800    \\ 
Exp. Spread  & Recomputed     &\textit{3.25e-01}    &       \textit{5.02e-01}    &        \textit{1.08e+00} \\
Exp. Confined  & Recomputed  & {\bf 4,600}     &   {\bf 5,900}   &
{\bf 7,500}         \\ 
Periodic & Recomputed &\textit{3.94e+02} \textit{(DNC)}& \textit{6.63e+02} & \textit{1.17e+03} \\ 
Periodic & Unchanged & \textit{8.91e+02} \textit{(DNC)} &
\textit{1.03e+03} &\textit{1.34e+03} \\ 
Random &  Recomputed &  \textit{1.97e+01} \textit{(DNC)}& \textit{3.37e+01} & \textit{8.10e+01}  \\ 
Random &  Unchanged & \textit{5.48e+02} \textit{(DNC)}&\textit{6.69e+02}  & \textit{9.65e+02}   \\ 
SetOnBoundary & Recomputed      &  900 (44) & 1,300 & 5,100       \\ 
SetOnBoundary & Reflected       & 242,100      & 387,100 &  811,400           \\   
SetOnBoundary & Set to Zero      &  1,300 (48)      & 1,900 & 4,100                 \\ 
Shrink & Recomputed       &  8,200 (49)      & 10,900 & 14,300                 \\ 
Shrink & Set to Zero      &  33,000      & 40,700 & 53,900                 \\ 
Hyperbolic & Modified (Eq.~(\ref{eq:hyperbolic}))		  & 14,100 &15,100 & 16,500			\\	\hline \hline
\multicolumn{5}{|c|}{	{$F_{\rm elp}$ in [-10,10]: At the Center}}	\\ \hline
IP Spread  & Recomputed            & 31,600   & 34,000   &37,900       \\ 
IP Confined  & Recomputed         & 30,900   & 33,800   & 38,500     \\ 
Exp. Spread  & Recomputed      & 30,500   & 34,700  & 38,300          \\ 
Exp. Confined   & Recomputed    & 31,900   & 35,100  & 38,200          \\ 
Periodic & Recomputed    & 32,200   & 35,100  & 37,900    \\ 
Periodic & Unchanged      & 33,800   & 36,600  & 41,200          \\
Random  & Recomputed        & 31,900   & 34,800 &  37,400      \\ 
Random & Unchanged          & 31,600   & 34,900    & 38,100             \\ 
SetOnBoundary &Recomputed  & 31,900   & 35,500    & 40,500       \\    
SetOnBoundary & Reflected   & 50,800 (38)  & 83,200  & 484,100        \\ 
SetOnBoundary & Set to Zero    & 31,600   & 35,000     & 37,200        \\ 
Shrink &  Recomputed           &  32,000  & 34,400 & 48,200                 \\
Shrink & Set to Zero             &  31,400  & 34,000 & 37,700                 \\ 
Hyperbolic & Modified (Eq.~(\ref{eq:hyperbolic}))
& {\bf 29,400} &  {\bf 31,200}  & {\bf 34,700}			\\ \hline \hline
\multicolumn{5}{|c|}{	  {$F_{\rm elp}$ in [-1,10]: Close to Boundary}}        \\ \hline
IP Spread  & Recomputed            &  28,200       & 31,900    & 35,300    \\ 
IP Confined  & Recomputed       &  28,300  & 32,900    & 44,600     \\ 
Exp. Spread  & Recomputed     &  28,300      &  30,700  & 33,200     \\
Exp. Confined & Recomputed     &  29,500      &  33,000  & 44,700     \\ 
Periodic & Recomputed       &  \textit{4.86e+01} \textit{(DNC)}  &   \textit{1.41e+02}  &  
\textit{4.28e+02}   \\
Periodic & Unchanged &  \textit{2.84e+02}  \textit{(DNC)} & \textit{5.46e+02}  & \textit{8.28e+02} \\
Random &  Recomputed         &  36,900      &  41,900  & 45,600     \\ 
Random & Unchanged &  \textit{1.13e+02}  \textit{(DNC)} & \textit{2.26e+02} & \textit{4.35e+02} \\ 
SetOnBoundary & Recomputed  &  \textit{1.80e+01} \textit{(DNC)} & \textit{7.60e+01} &  
\textit{3.00e+02}  \\ 
SetOnBoundary & Reflected   &  \textit{2.13e-01} \textit{(DNC)}& \textit{2.17e+01} &  
	\textit{1.06e+02}  \\
SetOnBoundary &  Set to Zero    &  31,700 (2)     &  31,700  & 32,600     \\ 
Shrink &  Recomputed           &  29,500 (6)      &  36,100  & 42,300     \\ 
Shrink &  Set to Zero             & 28,400 (36)      &  32,700  & 65,600     \\ 
Hyperbolic & Modified (Eq.~(\ref{eq:hyperbolic}))
& {\bf 25,900} & {\bf 29,200} & {\bf 31,000}	\\ \hline
\end{tabular}
\end{center}
\label{tab:PSOEllp}
\end{minipage}
\end{footnotesize}
\end{table*}
\begin{table*}
\begin{footnotesize}
\begin{minipage}[b]{1.0\linewidth}
\begin{center}
\caption{Results on $F_{\rm sch}$ with PSO for $10^{-10}$ termination criterion.}
\begin{tabular}{|lrrrr|} \hline \hline
{{Strategy}}  & Velocity   & {{Best}}& {{Median}}& {{Worst}} \\ \hline \hline
\multicolumn{5}{|c|}{{$F_{\rm sch}$ in [0,10]: On the Boundary}} \\ \hline
IP Spread   &  Recomputed         &  67,200  & 257,800 & 970,400       \\
IP Confined   &  Recomputed          &  112,400 (6)  & 126,500 & 145,900       \\
Exp. Spread    &  Recomputed     &  \textit{3.79e+00} \textit{(DNC)}& \textit{8.37e+00} & \textit{1.49e+01}  \\ 
Exp. Confined  &  Recomputed      & {\bf  4,900} & {\bf 6,100}   & {\bf  13,500}        \\
Periodic & Recomputed      &  \textit{4.85e+03} \textit{(DNC)}& \textit{7.82e+03}&\textit{1.34e+04}  \\ 
Periodic &Unchanged & \textit{7.69e+03} \textit{(DNC)}& \textit{1.11e+04}& \textit{1.51e+04}    \\
Random &Recomputed    &  \textit{2.61e+02} \textit{(DNC)}& \textit{5.44e+02}  &  \textit{1.05e+03}       \\ 
Random &Unchanged   &  \textit{5.30e+03} \textit{(DNC)} & \textit{7.60e+03} & \textit{1.22e+04} \\
SetOnBoundary &Recomputed  & 800 (30) & 1,100     &   3,900     \\  
SetOnBoundary &Reflected    & 171,500  & 241,700   &   434,200     \\ 
SetOnBoundary &Set to Zero   & 1,000 (40)      &  1,600      &  5,300              \\
Shrink &Recomputed	        &   6,900    &  9,100     &     11,600                           \\
Shrink &Set to Zero             &  17,900     &   31,900    &    49,800                            \\ 
Hyperbolic & Modified (Eq.~(\ref{eq:hyperbolic}))			& 36,400 & 41,700 & 48,700			\\   \hline   \hline
\multicolumn{5}{|c|}{{$F_{\rm sch}$ in [-10,10]: At the Center}} \\ \hline
IP Spread  &  Recomputed            & 106,700 & 127,500 & 144,300       \\ 
IP Confined  &  Recomputed          &  111,500  & 130,100    & 149,900     \\
Exp. Spread  &  Recomputed      &  112,300  & 131,400   &  149,000         \\ 
Exp. Confined  &  Recomputed  &  116,400  & 131,300   &  148,200         \\ 
Periodic &Recomputed        &  113,400  & 130,900    &    150,600  \\
Periodic &Unchanged         & 121,200   & 137,800       & 159,100          \\ 
Random &Recomputed          &  112,900  & 129,800     &  151,100       \\ 
Random &Unchanged           & 117,000  &   130,600    &  148,100        \\ 
SetOnBoundary &Recomputed   &  118,500 (49)  & 132,300       & 161,100       \\ 
SetOnBoundary &Reflected    & \textit{3.30e-06} \textit{(DNC)}& \textit{8.32e+01}\textit{(DNC)}&         
\textit{2.95e+02} \textit{(DNC)}   \\
SetOnBoundary & Set to Zero    &  111,900  & 132,200      &   149,700               \\ 
Shrink.&Recomputed            & 111,800 (49)	& 131,800	& 183,500  \\ 
Shrink.&Set to Zero             & 108,400	& 125,100	& 143,600  \\ 
Hyperbolic & Modified (Eq.~(\ref{eq:hyperbolic}))			& {\bf 101,300}	& 	{\bf  117,700} & {\bf 129,700} \\\hline \hline
\multicolumn{5}{|c|}{{$F_{\rm sch}$ in [-1,10]: Close to
  Boundary}} \\ \hline
IP Spread   &  Recomputed           &  107,200         &  130,400       &  272,400      \\ 
IP Confined  &  Recomputed         &  120,100 (44)        & 171,200        & 301,200    \\ 
Exp. Spread  &  Recomputed       &   92,800   & 109,200        &  126,400        \\ 
Exp. Confined &  Recomputed     &   110,200   & 127,400        &  256,100        \\ 
Periodic&Recomputed        &    \textit{8.09e+02} \textit{(DNC)}& \textit{2.01e+03} \textit{(DNC)}& 
\textit{5.53e+03}\textit{(DNC)}      \\ 
Periodic&Unchanged         &     \textit{2.16e+03} \textit{(DNC)}       &      \textit{4.36e+03} \textit{(DNC)}     &       \textit{6.87e+03} \textit{(DNC)}          \\ 
Random&Recomputed          & 123,300 & 165,600	& 280,000         \\ 
Random&Unchanged           &   \textit{8.17e+02} \textit{(DNC)} & \textit{1.96e+03} \textit{(DNC)}  &   
\textit{2.68e+03} \textit{(DNC)}               \\ 
SetOnBoundary&Recomputed   &    \textit{2.50e+00} \textit{(DNC)}   &   \textit{1.25e+01} \textit{(DNC)}   &         \textit{5.75e+02} \textit{(DNC)}       \\ 
SetOnBoundary&Reflected    &     \textit{1.86e+00} \textit{(DNC)}        &      \textit{7.76e+00} \textit{(DNC)}      &       \textit{5.18e+01} \textit{(DNC)}          \\ 
SetOnBoundary&Set to Zero    &   \textit{1.00e+00} \textit{(DNC)}     &   \textit{5.00e+00} \textit{(DNC)}     &       
\textit{4.21e+02}  \textit{(DNC)}              \\ 
Shrink & Recomputed            &      \textit{5.00e-01} \textit{(DNC)} &       \textit{3.00e+00} \textit{(DNC)}       &     \textit{1.60e+01} \textit{(DNC)}       \\ 
Shrink &Set to Zero             &   108,300 (8)   & 130,300        &  143,000        \\ 
Hyperbolic 	& Modified (Eq.~(\ref{eq:hyperbolic}))		& {\bf 93,100} 		&{\bf 	108,300}   &  {\bf 119,000}	\\ \hline
\end{tabular}
\label{tab:PSOSch}
\end{center}
\end{minipage}
\end{footnotesize}
\end{table*}

\begin{table*}[ht]
\begin{footnotesize}
\begin{minipage}[b]{1.0\linewidth}
\caption{Results on $F_{\rm ack}$ with PSO for $10^{-10}$ termination criterion.  }
\begin{center}
\begin{tabular}{|lrrrr|} \hline 
{{Strategy}}     & {{Best}}& {{Median}}& {{Worst}} \\ \hline \hline
\multicolumn{5}{|c|}{{$F_{\rm ack}$ in [0,10]: On the Boundary}} \\ \hline
IP Spread  &  Recomputed		&	150,600 (49)	& 220,900	 & 328,000  \\ 
IP Confined &  Recomputed	& \textit{4.17e+00} \textit{(DNC)}  & \textit{6.53e+00} & \textit{8.79e+00}\\ 
Exp. Spread &  Recomputed	&\textit{2.76e-01} \textit{(DNC)}  & \textit{9.62e-01}  & 
\textit{2.50e+00} \\ 
Exp. Confined &  Recomputed	& 7,800	& 9,600	& 11,100 \\
Periodic&Recomputed  	&\textit{6.17e+00} \textit{(DNC)} & \textit{6.89e+00}  & 
\textit{9.22e+00}  \\ 
Periodic&Unchanged 	 & \textit{8.23e+00} \textit{(DNC)}  & \textit{9.10e+00} & 
\textit{9.68e+00} \\ 
Random&Recomputed  	& \textit{3.29e+00} \textit{(DNC)} & \textit{3.40e+00} & \textit{4.19e+00} \\
Random&Unchanged &\textit{6.70e+00} \textit{(DNC)}& \textit{7.46e+00}& \textit{8.57e+00} 
\\ 
SetOnBoundary&Recomputed  	&{\bf  800}	& {\bf 1,100}	& {\bf 2,100}  \\
SetOnBoundary&Reflected 	 & 420,600	& 598,600	& 917,400  \\ 
SetOnBoundary&Set to Zero	 & 1,100	& 1,800	& 3,100	 \\
Shrink.&Recomputed  		& 33,800 (5) & 263,100  & 690,400  \\
Shrink.&Set Zero 		& \textit{3.65e+00} \textit{(DNC)}  & \textit{6.28e+00}  & \textit{8.35e+00} \\ 
Hyperbolic & Modified (Eq.~(\ref{eq:hyperbolic}))		& 24,600 (25) &26,100 &28,000 \\	\hline \hline
\multicolumn{5}{|c|}{   {$F_{\rm ack}$ in [-10,10]: At the Center}} \\ \hline
IP Spread &  Recomputed			 &	53,900 (46)	& 58,600	& 66,500  \\
IP Confined &  Recomputed			& 54,800 (49)	& 59,200	& 64,700  \\ 
Exp. Spread &  Recomputed		& {\bf 55,100}	& {\bf 59,300}	& {\bf 63,600} \\ 
Exp. Confined	&  Recomputed	 &	 56,800 & 59,600	& 65,000  \\ 
Periodic&Recomputed 		 & 55,700 (48)	& 59,900	& 64,700  \\ 
Periodic&Unchanged 		 &	57,900 (49)	& 62,100	& 66,700  \\
Random&Recomputed  		& 55,100 (47)	& 59,400	& 65,100  \\ 
Random&Unchanged 		&	56,300	 & 59,700	& 65,500  \\
SetOnBoundary&Recomputed 	 & 55,100 (49)	& 58,900	& 65,400 \\ 
SetOnBoundary&Reflected  	& 86,900 (4)	& 136,400	& 927,600 \\
SetOnBoundary&Set to Zero 	& 53,900 (49)	& 59,600	& 67,700 \\ 
Shrink &Recomputed  		& 55,800 (47)		& 58,700	& 65,800 \\ 
Shrink &Set to Zero 			& 55,700 (49)		& 58,900	& 62,000  \\ 
Hyperbolic & Modified (Eq.~(\ref{eq:hyperbolic}))			& 52,900 (49) & 56,200 & 64,400		\\\hline \hline
\multicolumn{5}{|c|}{   {$F_{\rm ack}$ in [-1,10]: Close to
  Boundary}} \\ \hline
IP Spread &  Recomputed			 &	54,600 (5)	& 55,100	& 56,600 \\ 
IP Confined  &  Recomputed			& 63,200 (1)	& 63,200	& 63,200  \\ 
Exp. Spread &  Recomputed		& {\bf 51,300}	& {\bf 55,200}	& {\bf 58,600} \\ 
Exp. Confined &  Recomputed & \textit{1.42e+00} \textit{(DNC)} & \textit{2.17e+00} 
& \textit{2.92e+00}  \\
Periodic&Recomputed  & \textit{2.88e+00} \textit{(DNC)}  & \textit{4.03e+00} & \textit{5.40e+00}  \\
Periodic&Unchanged  & \textit{6.61e+00} \textit{(DNC)}  & \textit{7.46e+00}  & \textit{8.37e+00} \\ 
Random&Recomputed  & 60,300 (45) & 66,200 & 72,200 \\
Random&Unchanged  	& \textit{4.21e+00} \textit{(DNC)} &
\textit{4.93e+00}  & \textit{6.11e+00} \\ 
SetOnBoundary&Recomputed & \textit{2.74e+00} \textit{(DNC)} & \textit{3.16e+00} & \textit{3.36e+00} \\
SetOnBoundary&Reflected 	 & 824,700 (1)	& 824,700	& 824,700 \\
SetOnBoundary&Set to Zero 	& \textit{1.70e+00} \textit{(DNC)} & \textit{2.63e+00} 
& \textit{3.26e+00}  \\ 
Shrink&Recomputed  & \textit{1.45e+00} \textit{(DNC)} & \textit{2.34e+00}  & \textit{2.73e+00} \\ 
Shrink&Set to Zero & \textit{2.01e+00} \textit{(DNC)} & \textit{3.96e+00}  & \textit{6.76e+00}  \\ 
	Hyperbolic & Modified (Eq.~(\ref{eq:hyperbolic})) &
        50,000 (39) &  53,500 & 58,100 \\ \hline
\end{tabular}
\label{tab:PSOAck}
\end{center}
\end{minipage}
\end{footnotesize}
\end{table*}

\begin{table*}[ht]
\begin{footnotesize}
\caption{Results on $F_{\rm ros}$ with PSO for $10^{-10}$ termination criterion.}
\begin{minipage}[b]{1.0\linewidth}
\begin{center}
\begin{tabular}{|lrrrr|} \hline 
{{Strategy}}     & {{Best}}& {{Median}}& {{Worst}} \\ \hline \hline
\multicolumn{5}{|c|}{   {$F_{\rm ros}$ in [1,10]: On the Boundary}} \\ \hline
IP Spread  &  Recomputed                   & 89,800 & 195,900  & 243,300	 \\ 
IP Confined &  Recomputed                  & 23,800 &164,300 & 209,300	\\ 
Exp. Spread  &  Recomputed   & \textit{9.55e-01} \textit{(DNC)} & \textit{2.58e+00}
& \textit{7.64e+00} \\ 
Exp. Confined  &  Recomputed            & {\bf 3,700} &	{\bf 128,100}	& {\bf 344,400}	\\ 
Periodic&Recomputed                & \textit{1.24e+04} \textit{(DNC)}  & \textit{2.35e+04}
	&\textit{4.24e+04} \\
Periodic&Unchanged                 & \textit{6.99e+04}  \textit{(DNC)} & \textit{1.01e+05} 
& \textit{1.45e+05} 		\\
Random&Recomputed           & \textit{6.00e+01} \textit{(DNC)}  & \textit{1.37e+02} & 
	\textit{4.42e+02}  		  \\
Random&Unchanged          & \textit{2.32e+04} \textit{(DNC)}  & \textit{3.90e+04} 
& \textit{8.22e+04}  \\ 
SetOnBoundary&Recomputed    	&  900 (45) &	1,600 	&89,800		 \\ 
SetOnBoundary&Reflected     	&  \textit{2.14e-03} \textit{(DNC)}  & \textit{6.01e+02} 
& \textit{5.10e+04} 		 \\ 
SetOnBoundary&Set to Zero    	& 1,400 (48) &	3,000	& 303,700 		 \\ 
Shrink.&Recomputed             	& 3,900 (44) &	5,100 	& 406,000		 \\
Shrink.&Set to Zero                     & 15,500 &136,200 & 193400		 \\
Hyperbolic    &   & 177,400 (45) & 714,300 & 987,500\\		 \hline \hline
\multicolumn{5}{|c|}{   {$F_{\rm ros}$ in [-8,10]: Near the
  Center}} \\ \hline
IP Spread   &  Recomputed                    & 302,300 (28) & 774,900 & 995,000 	   \\
IP Confined  &  Recomputed                  & 296,600 (32) &729,000 &955,000 	 \\ 
Exp. Spread  &  Recomputed              &  208,800 (24) & 754,700 & 985,200  	 \\ 
Exp. Confined &  Recomputed              &  301,100 (33) & 801,400 & 961,800      \\ 
Periodic&Recomputed                &  26,200 (27)  & 705,100  & 986,200        \\
Periodic&Unchanged                 &  247,300 (32) & 776,800 & 994,900  	    \\
Random&Recomputed                  &  311,200 (30)  &809,300 & 990,800        \\ 
Random&Unchanged                   & 380,100 (29) & 793,300 & 968,300	 \\
SetOnBoundary&Recomputed           & {\bf 248,700} (35) & {\bf 795,600} & {\bf 973,900}  \\ 
SetOnBoundary&Reflected            &  661,900 (01) & 661,900  & 661,900      \\
SetOnBoundary&Set to Zero           & 117,400 (25) &	858,400 	& 995,400              \\
Shrink.&Recomputed                    & 347,900 (33) & 790,500& 996,300                \\
Shrink.&Set to Zero                     & 353,300 (26) & 788,700 & 986,800              \\ 
Hyperbolic   & Modified (Eq.~(\ref{eq:hyperbolic}))	           & \textit{6.47e-08 (DNC)} & \textit{1.27e-04 (DNC)}& \textit{6.78e+00 (DNC)} 	\\	\hline \hline
\multicolumn{5}{|c|}{   {$F_{\rm ros}$ in [1,10]: Close to Boundary}}
\\ \hline
IP Spread  &  Recomputed                    & 184,600 (47) &  442,200  & 767,500       \\
IP Confined &  Recomputed                   & 229,900 (40) & 457,600 & 899,200          \\
Exp. Spread  &  Recomputed               &  19,400 (47) & 378,200 & 537,300       \\ 
Exp. Confined &  Recomputed             &  \textit{6.79e-03} \textit{(DNC)}   & \textit{4.23e+00} & 
\textit{6.73e+01}   \\ 
Periodic&Recomputed                &  \textit{1.51e-02} \textit{(DNC)}  	& \textit{3.73e+00} 
& \textit{5.17e+02}      \\
Periodic&Unchanged             &  \textit{1.92e+04} \textit{(DNC)}  & \textit{2.86e+04} & 
\textit{6.71e+04}       \\ 
Random&Recomputed                  & {\bf 103,800} 	& {\bf 432,200}	& {\bf 527,200}      \\ 
Random&Unchanged                   &   \textit{2.33e+02} \textit{(DNC)}  & \textit{1.47e+03}  
& \textit{4.23e+03}  \\  
SetOnBoundary&Recomputed           & \textit{1.71e+01} \textit{(DNC)} & \textit{1.87e+01}& 
\textit{3.13e+02}  	\\
SetOnBoundary&Reflected            & \textit{6.88e+00} \textit{(DNC)}  	& \textit{5.52e+02}  & 
	\textit{2.14e+04}  \\
SetOnBoundary&Set to Zero            & \textit{6.23e+00} \textit{(DNC)} & \textit{1.80e+01}
& \textit{3.12e+02}          \\ 
Shrink &Recomputed                    & 350,300 (3) 	& 350,900	& 458,400               \\ 
Shrink &Set to Zero                     & 163,700 (26) & 418,000  &531,900   \\ 
Hyperbolic & Modified (Eq.~(\ref{eq:hyperbolic}))	 & 920,900 (1) & 920,900 & 920,900     	\\ 	\hline
\end{tabular}
\label{tab:PSORos}
\end{center}
\end{minipage}
\end{footnotesize}
\end{table*}

\begin{figure}[hbt]
\begin{center}
\includegraphics[scale=1.0]{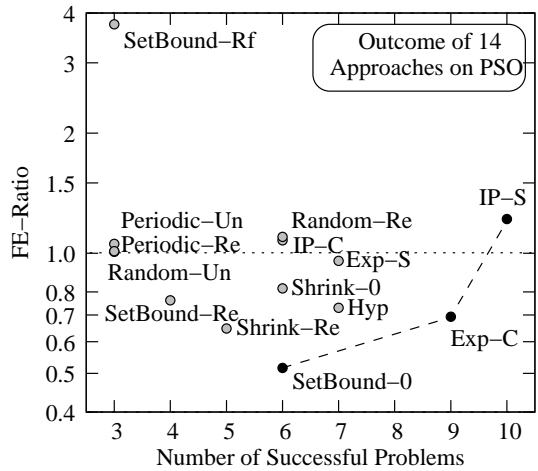}    
\end{center}
\caption{Performance comparison of 14 approaches for two metrics --
  number of problems solved successfully and function evaluation ratio
  -- with the PSO algorithm.}
\label{fig:pso_rank}
\end{figure}

The extensive simulation results are summarized using the following method. For each (say $j$) of the 14
approaches, the corresponding number of the successful applications ($\rho_j$) are recorded. Here,
an application is considered to be successful if more than 45 runs out of 50
runs are able to find the optimum within the specified accuracy. It is
observed that IP-S is successful in 10 out of 12 problem instances. Exponential confined approach (Exp-C) is successful in 9
cases. To investigate the required number of function evaluations (FE) needed
to find the optimum, by an approach (say $j$), we compute the average number of $\bar{\rm FE}_k^j$ 
needed to solve a particular problem ($k$) and construct the following objective for
$j$-th approach:
\begin{equation}
\mbox{FE-ratio}_j = \frac{1}{\rho_j}\sum_{k=1}^{12} \frac{\mbox{FE}_k^j}{\bar{\rm FE}_k^j}, 
\end{equation}
where FE$_k^j$ is the FEs needed by the $j$-th approach to solve the
$k$-th problem. Figure~\ref{fig:pso_rank} shows the performance of
each ($j$-th) of the 14 approaches on the two-axes plot ($\rho_j$ and
$\mbox{FE-ratio}_j$). 

The best approaches should have large $\rho_j$ values and small $\mbox{FE-ratio}_j$ values. 
This results in a trade-off between three best approaches which are marked in filled circles. All other 11
approaches are {\em dominated\/} by these three approaches. The
{\it SetBound} (\textit{SetOnBoundary}) with velocity set to zero performs in only six out of 12
problem instances. Thus, we ignore this approach. There is a clear
trade-off between IP-S and Exp-C approaches. IP-S solves one
problem more, but requires more FEs in comparison to Exp-C. Hence, we
recommend the use of both of these methods vis-a-vis all other methods used in this study. 

Other conclusions of this extensive study of PSO with different
constraint-handling methods are summarized as follows:
\begin{enumerate}
\item The constraint-handling methods show a large variation in the performance
depending on the choice of test problem and location of the optimum in
the allowable variable range.

\item When the optimum is on the variable boundary, periodic and
  random allocation methods perform poorly. This
  is expected intuitively. 

\item When the optimum is on the variable boundary, methods that set infeasible
  solutions on the violated boundary (\textit{SetOnBoundary} methods) work very
  well for obvious reasons, but these methods do not perform well for other cases. 

\item When the optimum lies near the center of the allowable range,
  most constraint-handling approaches work almost equally well. This can be understood intuitively from the fact that
tendency of particles to fly out of the search space is small when the optimum is
in the center of the allowable range. For example, the
periodic approaches fail in all the cases but are able to demonstrate
some convergence characteristics 
for all test problems, when the optimum is at the center. When the optimum is on the boundary or close
to the boundary, then the effect of the chosen 
constraint-handling method becomes critical.

\item The shrink method (with ``Velocity Recomputed" and ``Velocity Set
  Zero'' strategies) succeeded in 10 of the 12 cases.
\end{enumerate}

\subsection{Parametric Study of $\alpha$}\label{sec:alpha-effect}
The proposed IP approaches involve a parameter $\alpha$ affecting the
variance of the probability distribution for the mapped variable. In
this section, we perform a parametric study of $\alpha$ to determine
its effect on the performance of the IP-S approach.
 
\begin{figure}[hbt]
\begin{center}
\includegraphics[scale=.75]{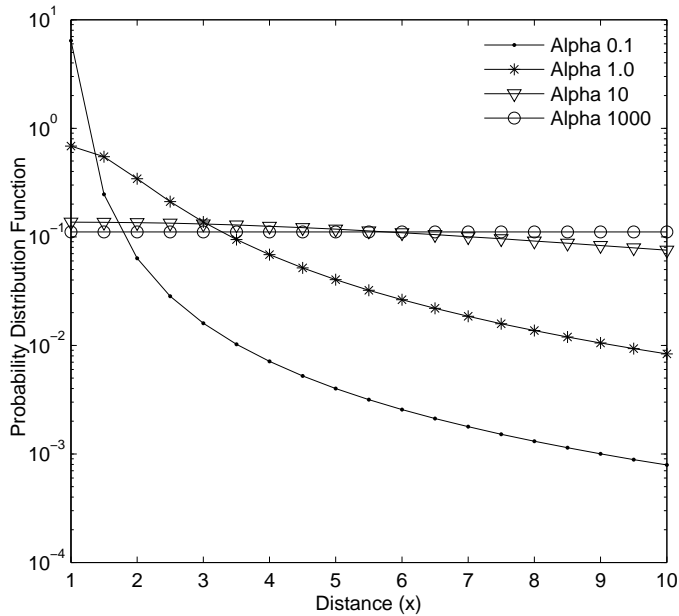}    
\caption{Probability distribution function with different $\alpha$
  values. The $x$-axis denotes the increased distance from the
  violated boundary. $x=1$ means the violated boundary. The child is
  created at $x=0$.}
\label{fig:Alpha-effect-figure}
\end{center}
\end{figure}

Following $\alpha$ values are chosen: $0.1$, $1$, $10$, and
$1,000$. To have an idea of the effect of $\alpha$, we plot the
probability distribution of mapped values in the allowable range
($[1,10]: On the Boundary$) for $d=1.0$ in Figure~\ref{fig:Alpha-effect-figure}. It can
be seen that for $\alpha=10$ and 1,000, the distribution is almost
uniform. 

\begin{figure}[hbt]
\begin{center}
\includegraphics[scale=1.0]{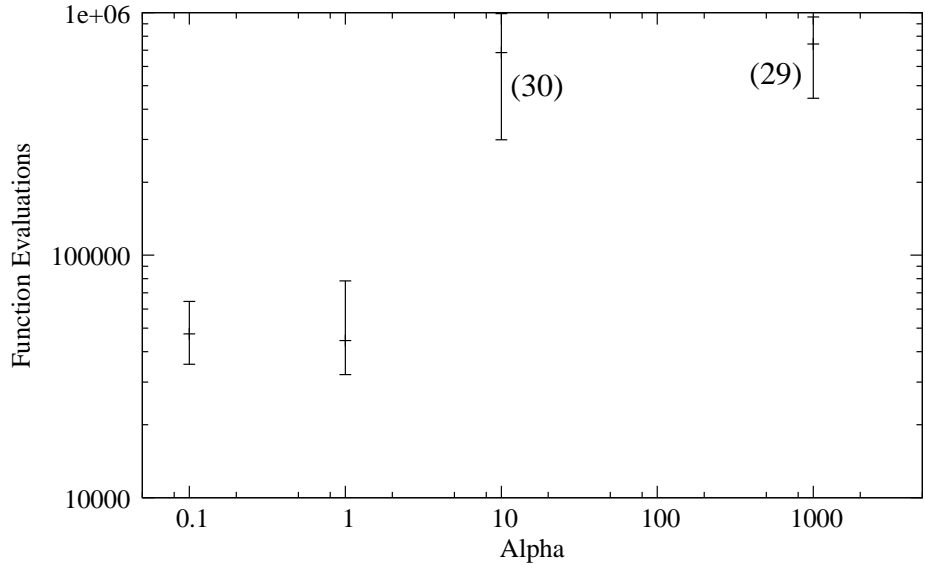}    
\caption{Performance of PSO algorithm with the IP-S approach with
  different $\alpha$ values on $F_{elp}$ problem.}
\label{fig:Alpha-effect-result}
\end{center}
\end{figure}

Figure ~\ref{fig:Alpha-effect-result} shows the effect
of $\alpha$ on $F_{elp}$ problem. For the same termination criterion 
we find that $\alpha=0.1$ and $1.0$ perform better compared to other
values. With
larger values of $\alpha$ the IP-S method does not even find the
desired solution in all 50 runs. 

\subsection{Results with Differential Evolution (DE)}\label{subsec:DEresults}
Differential evolution, originally proposed in \cite{storn}, has gained 
popularity as an efficient evolutionary optimization algorithm. 
The developers of DE proposed a total of ten different strategies 
\cite{StornPriceBook}. In \cite{DE-Boundary-Handling} it was shown that 
performance of DE largely depended upon the choice of constraint-handling mechanism.
We use Strategy~1 (where the offspring is created around the population-best solution), which is
most suited for solving unimodal problems \cite{padhyeJOGO2012}. 
A population size of $50$ was chosen with parameter values of  
$CR=0.5$ and $F=0.7$. Other parameters are set the same as before. 
We use $S=10^{-10}$ as our termination criterion. Results are tabulated 
in Tables~\ref{tab:DEellp} to ~\ref{tab:DEros}. 
Following two observations can be drawn:
\begin{enumerate}
\item For problems having optimum at one of the boundaries,
  {\it SetOnBoundary} approach performs the best. This is not a surprising
  result.
\item However, for problems having the optimum near the center of the
  allowable range, almost all eight algorithms perform in a similar
  manner.
\item For problems having their optimum close to one of the boundaries,
the proposed IP and existing exponential approaches perform better than
the rest of the approaches with DE.
\end{enumerate}
Despite the differences, somewhat similar performances of different constraint-handling
approaches with DE indicates
that the DE is an efficient optimization algorithm and its performance
is somewhat less dependent on the choice of constraint-handling scheme
compared to the PSO algorithm.
\begin{table*}[ht]
\begin{footnotesize}
\begin{center}
\caption{Results on $F_{\rm elp}$ with DE for $10^{-10}$ termination criterion.}
\begin{tabular}{|lrrr|} \hline 
{{Strategy}}     & {{Best}}& {{Median}}& {{Worst}} \\ \hline \hline
\multicolumn{4}{|c|}{$F_{\rm elp}$ in [0,10]: On the Boundary} \\ \hline
IP Spread       &       25,600   & 26,850 &27,650  \\  
IP Confined   &       22,400   &23,550  &24,200  \\  
Exp. Spread  &      38,350   & 39,800 &41,500  \\  
Exp. Confined &     19,200   & 20,700 & 21,900 \\  
Periodic                &       42,400   & 43,700 &       45,050   \\ 
Random                  &       40,650   & 43,050 & 44,250 \\  
SetOnBoundary           & {\bf 2,850}         &     {\bf   3,350}    & {\bf 3,900} \\  
Shrink                    & 4,050          & 4,900          & 5,850  \\       \hline \hline
\multicolumn{4}{|c|}{$F_{\rm elp}$ in [-10,10]: At the Center} \\ \hline  
IP Spread       & 29,950 & 31,200 & 32,500     \\  
IP Confined     & {\bf 29,600}  & {\bf 31,200} & {\bf 32,400}      \\ 
Exp. Spread  & 29,950 &31,300 & 32,400      \\  
Exp. Confined &  30,500& 31,400       & 32,250    \\
Periodic                &   29,650 &     31,300 &         32,400       \\ 
Random                  &  30,000 &      31,200 &         31,250     \\  
SetOnBoundary           & 29,850 & 31,200 & 32,700    \\ 
Shrink                     & 30,300 & 31,250  &32,750  \\        \hline \hline
\multicolumn{4}{|c|}{$F_{\rm elp}$ in [-1,10]: Close to Boundary} \\ \hline
IP Spread       & 28,550 & 29,600 &       30,550   \\ 
IP Confined    & 28,500 & 29,500 & 30,650   \\
Exp. Spread & {\bf 28,050} & {\bf 28,900} & {\bf 29,850}   \\ 
Exp. Confined& 28,150 &      29,050   & 29,850   \\ 
Periodic                &  29,850 &      30,850 &         32,100      \\
Random                  &  28,900 &      30,200   & 31,000    \\ 
SetOnBoundary           &  28,650 &      29,600   & 30,500  \\  
Shrink                     & 28,800 & 29,900 & 31,200  \\        \hline \hline
\end{tabular}
\label{tab:DEellp}
\end{center}
\end{footnotesize}
\end{table*}
\begin{table*}[ht]
\begin{center}
\begin{footnotesize}
\caption{Results on $F_{\rm sch}$ with DE for $10^{-10}$ termination criterion.}
\begin{tabular}{|lrrr|} \hline 
{{Strategy}}     & {{Best}}& {{Median}}& {{Worst}} \\ \hline \hline
\multicolumn{4}{|c|}{   {$F_{\rm sch}$ in [0,10]: On the Boundary}} \\ \hline
IP Spread       &       26,600 &         27,400   & 28,000          \\ 
IP Confined    &       22,450 &         23,350   & 24,300 \\  
Exp. Spread  &      40,500 &         42,050   & 43,200   \\ 
Exp. Confined&      19,650 &         20,350   & 22,050           \\  
Periodic                &     44,700   & 46,300           & 48,250   \\  
Random                  & 43,850       & 45,150           &       47,000     \\
SetOnBoundary           &  {\bf 2,100  }     &{\bf  3,100  }          & {\bf       3,750 } \\ 
Shrink                     & 3,450        & 4,400            &       5,100      \\     \hline \hline
\multicolumn{4}{|c|}{   {$F_{\rm sch}$ in [-10,10]: At the Center}} \\ \hline
IP Spread      &      {\bf  258,750} & {\bf 281,650}  & {\bf 296,300}   \\ 
IP Confined   &      268,150 & 283,050  & 300,450  \\ 
Exp. Spread &     266,850 & 283,950  & 304,500  \\ 
Exp. Confined &     266,450 & 283,700  & 305,550  \\ 
Periodic                &      269,700 & 284,100  & 310,100  \\ 
Random                  &      263,300 & 282,600  & 306,250         \\
SetOnBoundary           &      267,750 & 284,550  & 298,850         \\ 
Shrink                     &      263,600 & 282,750  & 304,350         \\        \hline \hline
\multicolumn{4}{|c|}{   {$F_{\rm sch}$ in [-1,10]: Close to Boundary}} \\ \hline
IP Spread    &    {\bf  228,950} & {\bf  242,300}  & {\bf 255,700}  \\
IP Confined     &     232,200 &  243,900  & 263,400  \\ 
Exp. Spread  &    227,550 &  243,000  & 261,950   \\ 
Exp. Confined &   228,750  &  243,800  & 262,500    \\ 
Periodic                &    231,950   & 247,150  & 260,700    \\  
Random                  &    228,550   & 244,850  & 261,900     \\  
SetOnBoundary           &    237,100  &  255,750  & 266,400     \\  
Shrink                     &    234,000  &  253,250  & 275,550     \\        \hline
\end{tabular}
\label{tab:DEsch}
\end{footnotesize}
\end{center}
\end{table*}
\begin{table*}[ht]
\begin{center}
\begin{footnotesize}
\caption{Results on $F_{\rm ack}$ with DE for $10^{-10}$ termination criterion.}
\begin{tabular}{|lrrr|} \hline
{{Strategy}}     & {{Best}}& {{Median}}& {{Worst}} \\ \hline \hline
\multicolumn{4}{|c|}{   {$F_{\rm ack}$ in [0,10]: On the Boundary}} \\ \hline
IP Spread      & 43,400 &       44,950 & 45,950    \\  
IP Confined    & 37,300 &       38,700 & 40,350    \\ 
Exp. Spread Dist. & 66,300 &      69,250 & 71,300      \\ 
Exp. Confined Dist& 32,750 &      34,600 & 36,200  \\ 
Periodic                &  72,500 &      74,250 & 75,900      \\ 
Random                  & 70,650  &      73,000 & 74,750       \\ 
SetOnBoundary           & {\bf  2,550} &    {\bf    3,250}  &{\bf  3,950 }      \\  
Shrink                     &  3,500 & 4,700        & 5,300       \\        \hline \hline
\multicolumn{4}{|c|}{   {$F_{\rm ack}$ in [-10,10]: At the Center}} \\ \hline
IP Spread     &     {\bf   50,650}   &{\bf  52,050} &     {\bf   53,450}     \\ 
IP Confined    &       51,050   & 52,200 &       53,800    \\  
Exp. Spread  &      51,200    &52,150  & 53,400    \\
Exp. Confined&      51,100   &52,300  & 53,850  \\  
Periodic                &       51,250   & 52,250  &53,500 \\  
Random                  &       50,950   & 52,200 & 53,450    \\ 
SetOnBoundary           &       50,950   & 52,300 &       53,450  \\ 
Shrink                     &      50,450    & 52,300 &       53,550   \\        \hline \hline
\multicolumn{4}{|c|}{   {$F_{\rm ack}$ in [-1,10]: Close to Boundary}} \\ \hline
IP Spread      &       49,100   & 50,650 & 51,650      \\  
IP Confined    &       48,650   & 50,400 & 52,100         \\
Exp. Spread &     {\bf  48,300}    & {\bf 49,900} & {\bf 51,750}   \\ 
Exp. Confined&     48,900    & 50,000 & 51,250 \\  
Periodic                &      50,400    & 51,950 & 53,300 \\  
Random                  &      50,250     &51,200  &52,150   \\ 
SetOnBoundary           &     49,900 (33) & 51,100 & 53,150  \\  
Shrink                     &     50,200     & 51,400 & 52,750  \\        \hline
\end{tabular}
\label{tab:DEack}
\end{footnotesize}
\end{center}
\end{table*}
\begin{table*}[ht]
\begin{center}
\begin{footnotesize}
\caption{Results on $F_{\rm ros}$ with DE for $10^{-10}$ termination criterion.}
\begin{tabular}{|lrrr|} \hline 
{{Strategy}}     & {{Best}}& {{Median}}& {{Worst}} \\ \hline \hline
\multicolumn{4}{|c|}{   {$F_{\rm ros}$ in [1,10]: On the Boundary}} \\ \hline
IP Spread       &       38,850   & 62,000 & 89,700 \\  
IP Confined    &       24,850   & 45,700 & 73,400               \\  
Exp. Spread  &      57,100   & 86,800 & 118,600   \\ 
Exp. Confined &     16,600    & 21,400  &79,550 \\  
Periodic                &    69,550      & 93,500 & 18,1150        \\ 
Random                  &    65,850       &92,950  &157,600     \\
SetOnBoundary           &   {\bf  2,950}        & {\bf 4,700} & {\bf 30,450}   \\  
Shrink                     &    5,450       & 8,150   &55,550 \\        \hline \hline
\multicolumn{4}{|c|}{   {$F_{\rm ros}$ in [-8,10]: At the Center}} \\ \hline
IP Spread      &       133,350 (41)      & 887,250&       995,700          \\ 
IP Confined    &       712,500 (44)      & 854,800 &      991,400          \\ 
Exp. Spread   &      {390,700} (48)       & {866,150} &      {998,950}   \\ 
Exp. Confined &     138,550 (40)        &883,500 &      994,350  \\  
Periodic                &     764,650 (39) &      874,700 &        999,650      \\
Random                  &     {\bf 699,400} (49)  &     {\bf 885,450} &       {\bf  999,600}   \\  
SetOnBoundary           &     743,600 (38)  &     865,450& 995,500 \\ 
Shrink                     &    509,900 (40)  &      873,450 &        998,450 \\    \hline    \hline
\multicolumn{4}{|c|}{   {$F_{\rm ros}$ in [0,10]: Close to Boundary}} \\ \hline
IP Spread      &      36,850    & 78,700 &       949,700          \\ 
IP Confined    &       46,400 (46) &     95,900 &         891,450          \\ 
Exp. Spread  &     49,550 (49)  &     85,900 &         968,200   \\
Exp. Confined &     87,300 (43)  &     829,200  & 973,350  \\  
Periodic                &     {\bf  38,750}    &{\bf  62,200}         & {\bf 94,750}    \\ 
Random                  &      41,200    & 61,300 &       461,500  \\ 
SetOnBoundary           &    8.23E+00 (DNC)    & 1.62E+01 &     1.89E+01        \\
Shrink                     &  252,650 (9) &  837,700  & 985,750         \\        \hline
\end{tabular}
\label{tab:DEros}
\end{footnotesize}
\end{center}
\end{table*}
\subsection{Results with Real-Parameter Genetic Algorithms (RGAs)}
\label{subsec:GAresults}
We have used two real-parameter GAs in our study here:
\begin{enumerate}
\item {\em Standard-RGA\/} using the simulated binary crossover (SBX) \cite{debramb} operator
and the polynomial mutation operator
  \cite{debBookMO}. In this approach, variables are expressed as
  real numbers initialized within the allowable range of each
  variable. The SBX and polynomial mutation operators can create
infeasible solutions. Violated boundary, if any, is handled using one
of the approaches studied in this paper. Later we shall investigate a
rigid boundary implementation of these operators which ensures
creation of feasible solutions in every recombination and mutation operations.

\item {\em Elitist-RGA} in which two newly created offsprings are compared against the two
parents, and the best two out of these four 
solutions are retained as parents (thereby introducing elitism). Here, the 
offspring solutions are created using non-rigid
 versions of SBX and polynomial mutation operators. 
As before, we test eight different constraint-handling approaches and, later
explore a rigid boundary implementation of the operators in presence of 
elite preservation. 
\end{enumerate}

Parameters for RGAs are chosen as follows: population size of 100, 
crossover probability $p_{c}$=0.9, mutation probability $p_{m}$=0.05, distribution index for crossover $n_{dist.,c}$=2, 
distribution index for mutation $n_{dist.,m}$=100.
The results for the Standard-RGA are shown in Tables~\ref{tab:RGAellp} to ~\ref{tab:RGAros}
for four different test problems. Tables~\ref{tab:GA-elitist-elp}
to ~\ref{tab:GA-elitist-ros} show results using the Elitist-RGA.
Following key observations can be made:
\begin{enumerate}
\item For all the four test problems, Standard-RGA shows convergence
  \textit{only} in the situation when optima is on the boundary.  
           
\item Elitist-RGA shows good convergence on $F_{elp}$ when the optimum is on the boundary and,
only some convergence is noted when the optima is at the other locations.
For other three problems, convergence is only obtained when optimum is present on the boundary.     

\item Overall, the performance of Elite-RGA is comparable or slightly better compared to Standard-RGA.
\end{enumerate}

The \textit{Did Not Converge} cases can be explained on the fact that
the SBX operator has the property of creating solutions around
the parents; if parents are close to each other. This property is a likely cause of premature
convergence as the population gets closer to the optima. 
Furthermore the results suggest that the elitism implemented in this study
(parent-child comparison) is not quite effective. 
 
Although RGAs are able to locate the optima,
however, they are unable to fine-tune the optima due to undesired 
properties of the generation scheme. This emphasizes the fact that 
generation scheme is primarily 
responsible for creating good solutions, and
the constraint-handling methods cannot act as surrogate
for generating efficient solutions. 
Each step of the evolutionary 
search should be designed effectively in order to achieve overall success. 
On the other hand one could argue that strategies such as increasing the mutation rate (in order to promote diversity so as to avoid
pre-mature convergence) should be tried, however, creation of good and meaningful solutions
in the generation scheme is rather an important and a desired fundamental-feature.
   
As expected, when the optima is on the boundary \textit{SetOnBoundary} finds the optima most efficiently within a minimum number
of function evaluations. Like in PSO the performance of 
\textit{Exp. Confined} is better than 
\textit{Exp. Spread}.
\textit{Periodic} and \textit{Random} show comparable 
or slightly worse performances (these mechanisms don't have any preference
of creating solutions close to the boundary and actually promote spread of 
the population). 

\begin{table*}[ht]
\begin{footnotesize}
\begin{minipage}[b]{1.0\linewidth}
\begin{center}
\caption{Results on $F_{\rm elp}$ with Standard-RGA for termination criterion of $10^{-10}$.}
\begin{tabular}{|lrrr|} \hline 
{{Strategy}}     & {{Best}}& {{Median}}& {{Worst}} \\ \hline \hline
\multicolumn{4}{|c|}{   {$F_{\rm elp}$ in [0,10]}} \\ 
IP Spread                & 9,200 & 10,500  & 12,900                        \\
IP Confined            & 7,900 &  9,400 & 10,900     \\
Exp. Spread        & 103,100 (6) & 718,900 & 931,200   \\ 
Exp. Confined        & 4,500 & 5,700 & 7,000                \\
Periodic                         & 15,200 (1) & 15,200 & 15,200       \\
Random                           & 68,300 (12) & 314,700 & 939,800      \\
SetOnBoundary                    & {\bf 1,800} &  {\bf 2,400} & {\bf 2,800}               \\
Shrink                             & 3,700 & 5,100 & 6,600 \\   \hline \hline
\multicolumn{4}{|c|}{   {$F_{\rm elp}$ in [-10,10]}} \\
\multicolumn{4}{|c|}{   	{2.60e-02 \textit{(DNC)}}}            \\ \hline \hline
\multicolumn{4}{|c|}{   {$F_{\rm elp}$ in [-1,10]}} \\ 	
\multicolumn{4}{|c|}{	{\textit{1.02e-02 (DNC)}}}		\\ \hline
\end{tabular}
\label{tab:RGAellp}
\end{center}
\end{minipage}
\end{footnotesize}
\end{table*}

\begin{table*}[ht]
\begin{footnotesize}
\begin{minipage}[b]{1.0\linewidth}
\begin{center}
\caption{Results on $F_{\rm sch}$ with Standard-RGA for termination criterion of $10^{-10}$}
\begin{tabular}{|lrrr|} \hline
{{Strategy}}     & {{Best}}& {{Median}}& {{Worst}} \\ \hline \hline
\multicolumn{4}{|c|}{   {$F_{\rm sch}$ in [0,10]}} \\
IP Spread               & 6,800         &9,800 & 11,800    \\
IP Confined           & 6,400         & 8,200  & 10,300     \\ 
Exp. Spread         & 21,200 (47)    & 180,000 &      772,200   \\
Exp. Confined       & 4,300  &      5,500 &  6,300                 \\
Periodic                         & 14,800 (26) &  143,500  & 499,400           \\ 
Random                           & 8,700 (43) &   195,200 &        979,300     \\
SetOnBoundary                    & {\bf 1,800} &     {\bf  2,300} &  {\bf 2,900}               \\
Shrink.                             & 3,600  &      4,600 &  5,500                    \\      \hline \hline
\multicolumn{4}{|c|}{   {$F_{\rm sch}$ in [-10,10]}} \\ 
\multicolumn{4}{|c|}{   {	\textit{1.20e-01 (DNC)}}}            \\ \hline \hline
\multicolumn{4}{|c|}{   {$F_{\rm sch}$ in [-1,10]}} \\
\multicolumn{4}{|c|}{   {\textit{8.54e-02 (DNC)}}}            \\ \hline
\end{tabular}
\label{tab:RGAsch}
\end{center}
\end{minipage}
\end{footnotesize}
\end{table*}

\begin{table*}[ht]
\begin{footnotesize}
\begin{minipage}[b]{1.0\linewidth}
\begin{center}
\caption{Results on $F_{\rm ack}$ with Standard-RGA for termination criterion of $10^{-10}$}
\begin{tabular}{|lrrr|} \hline 
{{Strategy}}     & {{Best}}& {{Median}}& {{Worst}} \\ \hline \hline
\multicolumn{4}{|c|}{   {$F_{\rm ack}$ in [0,10]}} \\ 
IP Spread       & 12,100         &       22,600   & 43,400                 \\ 
IP Confined    & 9,800          &       13,200   & 16,400                 \\  
Exp. Spread &      58,100 (29) &     355,900  & 994,000                \\   
Exp. Confined&    6,300      & 9,100        & 11,900                 \\   
Periodic                  & 19,600 (46)   & 122,300         &870,200                \\    
Random          &35,700 (38)      & 229,200        & 989,500                        \\     
SetOnBoundary   & {\bf 1,800} &       {\bf  2,500}    & {\bf 3,100}                                          \\
Shrink              &4,200  &       5,700 &  8,600                          \\ \hline   \hline
\multicolumn{4}{|c|}{   {$F_{\rm ack}$ in [-10,10]}} \\ 
\multicolumn{4}{|c|}{   { \textit{7.76e-02(DNC)}}}               \\ \hline \hline
 
\multicolumn{4}{|c|}{   {$F_{\rm ack}$ in [-1,10]}} \\
\multicolumn{4}{|c|}{   {\textit{4.00e-02 (DNC)}}}               \\ \hline
 
\end{tabular}
\label{tab:RGAack}
\end{center}
\end{minipage}
\end{footnotesize}
\end{table*}
\begin{table*}
\begin{footnotesize}
\begin{minipage}[b]{1.0\linewidth}
\begin{center}
\caption{Results on $F_{\rm ros}$ with Standard-RGA for termination criterion of $10^{-10}$}
\begin{tabular}{|lrrr|} \hline 
{{Strategy}}     & {{Best}}& {{Median}}& {{Worst}} \\ \hline \hline
\multicolumn{4}{|c|}{   {$F_{\rm ros}$ in [1,10]: On the Boundary}} \\ 
IP Spread        & 12,400 (39)&   15,800&  20,000        \\ 
IP Confined      & 9,400 (39)&    11,800&  13,600                \\
Exp. Spread  &\textit{9.73e+00} \textit{(DNC)} & \textit{1.83e+00}&  \textit{2.43e+01}        \\
Exp. Confined & 6,000       &       6,900&   8,200       \\ 
Periodic                  & \textit{6.30E+01} \textit{(DNC)}&  \textit{4.92e+02}&   \textit{5.27e+04}     \\ 
Random                    & \textit{3.97e+02} \textit{(DNC)}    &  \textit{9.28e+02}&     \textit{1.50e+03} \\
SetOnBoundary             &  {\bf 1,900}     &    {\bf   2,700} &   {\bf 3,400}                   \\
Shrink                       &  4,100      &      5,200 &  6,500                  \\ \hline \hline
\multicolumn{4}{|c|}{   {$F_{\rm ros}$ in [-8,10]}} \\
\multicolumn{4}{|c|}{   { \textit{3.64e+00 (DNC)}}} \\  \hline \hline 
\multicolumn{4}{|c|}{   {$F_{\rm ros}$ in [1,10]: On the Boundary}} \\ 
\multicolumn{4}{|c|}{   {\textit{1.04e+01(DNC)}}} \\  \hline 
\end{tabular}
\label{tab:RGAros}
\end{center}
\end{minipage}
\end{footnotesize}
\end{table*}



\clearpage
\begin{table*}[ht]
\begin{footnotesize}
\begin{center}
\caption{Results on $F_{\rm elp}$ with Elite-RGA for $10^{-10}$ termination criterion.}
\begin{tabular}{|lrrr|} \hline
{{Strategy}}     & {{Best}}& {{Median}}& {{Worst}} \\ \hline \hline
\multicolumn{4}{|c|}{   {$F_{\rm elp}$ in [0,10]}} \\ 
IP Spread       &  6,600 &      8,000    &  9,600                                  \\
IP Confined    &  6,300 & 8,100 & 9,800                                  \\ 
Exp. Spread   &     4,800 & 6,900  & 8,300                             \\ 
Exp. Confined &    4,600 & 5,800 & 6,700                                         \\
Periodic                  &    6,500 & 8,800 & 11,500                                       \\ 
Random                    &     6,400 & 7,900 & 10,300                                  \\ 
SetOnBoundary             &   {\bf 2,200} & {\bf 2,600} & {\bf 3,500}                                       \\ 
Shrink                       &   4,000 & 5,200 & 6,700                                        \\ \hline \hline
\multicolumn{4}{|c|}{   {$F_{\rm elp}$ in [-10,10]}} \\ 
IP Spread        &    980,200 (1) & 980,200 & 980,200               \\
IP Confined    &    479,000 (1) & 479,000 & 479,000                           \\
Exp. Spread &   \textit{2.06e-01} \textit{(DNC)} & \textit{4.53e-01} 
& \textit{4.86e-01}                        \\ 
Exp. Confined &   954,400 (1)	 & 954,400 & 954,400                                   \\
Periodic                  &   \textit{1.55E-01} \textit{(DNC)}& \textit{2.48E-01} & \textit{2.36E-01}  \\ 
Random                    &   \textit{1.92E-01} \textit{(DNC)}& \textit{2.00E-01} & \textit{2.46E-01}    \\ 
SetOnBoundary             &   \textit{2.11E-01} \textit{(DNC)}&2.95E-01 & 1.94E-01                       \\
Shrink                       &   {\bf 530,900} (3) & {\bf 654,000} & {\bf 779,000}                \\ \hline \hline
\multicolumn{4}{|c|}{   {$F_{\rm elp}$ in [-1,10]}} \\ 
IP Spread        &   803,400 (5) & 886,100 & 947,600             \\ 
IP Confined      &   643,300 (2) & 643,300 &963,000                          \\
Exp. Spread  &  593,300 (3) & 628,900 & 863,500                       \\ 
Exp. Confined &  655,400 (3) & 940,500 & 946,700                                         \\
Periodic                  &  653,800 (3) & 842,900 &843,100                                  \\ 
Random                    &  498,500 (2) & 498,500 &815,500                           \\ 
SetOnBoundary             &  {\bf 593,800} (5) & {\bf 870,500} & {\bf 993,500}                                \\ 
Shrink                       &  781,000 (2) & 781,000 &928,300                                   \\ \hline
\end{tabular}
\label{tab:GA-elitist-elp}
\end{center}
\end{footnotesize}
\end{table*}
\begin{table*}[ht]
\begin{footnotesize}
\begin{center}
\caption{Results on $F_{\rm sch}$ with Elite-RGA for termination criterion of $10^{-10}$}
\begin{tabular}{|lrrr|} \hline \hline
{{Strategy}}     & {{Best}}& {{Median}}& {{Worst}} \\ \hline \hline
\multicolumn{4}{|c|}{   {$F_{\rm sch}$ in [0,10]}} \\ \hline
IP Spread        &   5,000 &  6,500 & 7,900                     \\
IP Confined      &   4,900 & 6,500 & 7,900                          \\
Exp. Spread  &  4,300 & 5,800 & 7,800                         \\ 
Exp. Confined &  4,300 & 4,900 & 5,600                                   \\ 
Periodic                  &  5,400 & 7,000 & 11,300                                 \\ 
Random                    &  5,300 & 6,600 & 8,500                            \\ 
SetOnBoundary             &  {\bf 1,600} & {\bf 2,200} & {\bf 2,600}                                   \\
Shrink                       &  3,100 & 4,200 & 5,400              \\ \hline \hline
\multicolumn{4}{|c|}{   {$F_{\rm sch}$ in [-10,10]}} \\ 
\multicolumn{4}{|c|}{   {\textit{8.12e-05(DNC)}}}            \\ \hline \hline
\multicolumn{4}{|c|}{   {$F_{\rm sch}$ in [-1,10]}} \\
\multicolumn{4}{|c|}{   { \textit{1.61e-01(DNC)}}}            \\ \hline 
\end{tabular}
\label{tab:GA-elitist-sch}
\end{center}
\end{footnotesize}
\end{table*}
\begin{table*}[ht]
\begin{footnotesize}
\begin{center}
\caption{Results on $F_{\rm ack}$ with Elite-RGA for termination criterion of $10^-{10}$}
\begin{tabular}{|lrrr|} \hline \hline
{{Strategy}}     & {{Best}}& {{Median}}& {{Worst}} \\ \hline \hline
\multicolumn{4}{|c|}{   {$F_{\rm ack}$ in [0,10]}} \\ \hline
IP Spread       &    6,300 &8,700 &46,500                         \\
IP Confined     &    6,800 &9,200 &32,000                       \\ 
Exp. Spread  &   5,600 &6,800 &8,700                       \\ 
Exp. Confined &   5,200 &7,800 &9,900                               \\ 
Periodic                  &   6,300 &9,300 &12,200                              \\ 
Random                    &   6,200 &8,300 &53,700                        \\ 
SetOnBoundary             &   {\bf 1,900} &{\bf 2,500} &{\bf 4,000}                               \\
Shrink                       &   3,900 &5,100 &7,700                           \\  \hline \hline
\multicolumn{4}{|c|}{   {$F_{\rm ack}$ in [-10,10]}} \\ 
\multicolumn{4}{|c|}{   {\textit{1.03e-01(DNC)}}}            \\ \hline
\multicolumn{4}{|c|}{   {$F_{\rm ack}$ in [-1,10]}} \\ 
\multicolumn{4}{|c|}{   {\textit{1.15e-00 (DNC)}}}            \\ \hline
\end{tabular}
\label{tab:GA-elitist-ack}
\end{center}
\end{footnotesize}
\end{table*}
 
\begin{table*}[ht]
\begin{footnotesize}
\begin{center}
\caption{Results on $F_{\rm ros}$ with Elite-RGA for termination criterion of $10^-{10}$}
\begin{tabular}{|lrrr|} \hline \hline
{{Strategy}}     & {{Best}}& {{Median}}& {{Worst}} \\ \hline \hline
\multicolumn{4}{|c|}{   {$F_{\rm ros}$ in [0,10]}} \\ \hline
IP Spread       &   9,900 (13) &12,500 &14,000                    \\ 
IP Confined      &   10,100 (12)& 12,100 &14,400                      \\ 
Exp. Spread   &  8,500 (10) &11,000 &15,400                    \\ 
Exp. Confined &  6,600 (30) &7,800 &8,900                       \\ 
Periodic                  &  9,500 (10) &13,300 &16,800                          \\ 
Random                    &  14,000 (3)& 15,300 &16,100                          \\ 
SetOnBoundary             &  {\bf 2,300} (44) & {\bf 3,200} & {\bf 4,500}                              \\ 
Shrink                       &  4,500 (32) &6,100 &8,100                           \\ \hline \hline
\multicolumn{4}{|c|}{   {$F_{\rm ros}$ in [-8,10]}} \\ 
\multicolumn{4}{|c|}{   {\textit{1.27e-00 (DNC)}}}            \\ \hline \hline
 
\multicolumn{4}{|c|}{   {$F_{\rm ros}$ in [1,10]: On the Boundary}} \\ 
\multicolumn{4}{|c|}{   {\textit{1.49e-00 (DNC)}}}            \\ \hline 
\end{tabular}
\label{tab:GA-elitist-ros}
\end{center}
\end{footnotesize}
\end{table*}

\subsubsection{RGAs with Rigid Boundary}\label{subsec:NonelitistGAresults}

\begin{table*}[htb]
\begin{footnotesize}
\caption{RGA with rigid boundary with termination criterion $10^{-10}$}
{\footnotesize\begin{center}
\begin{tabular}{|lrrr|} \hline 
		\multicolumn{4}{|c|}{ Optimum on the boundary   }			\\	\hline	\hline	
{{Strategy}}     & {{Best}}& {{Median}}& {{Worst}} \\ \hline \hline
$F_{elp}$		&	8,100 & 8,500 & 8,800		\\	
$F_{sch}$		&	7,800 & 8,100 & 8,300		\\	
$F_{ack}$		&	9,500 & 10,100 & 10,800		\\	
$F_{ros}$		&	10,100 (39) & 10,900 & 143,600		\\	\hline	\hline 
\multicolumn{4}{|c|}{ Optimum in the center}		\\	
\multicolumn{4}{|c|}{ {\textit{3.88e-02 (DNC)}}}               \\
\hline \hline   
\multicolumn{4}{|c|}{ Optimum close to the edge of the boundary  }      \\		
\multicolumn{4}{|c|}{ {\textit{9.44e-03 (DNC)}}}               \\
\hline \hline   
\end{tabular}
\label{tab:rga-standard-rigid-boundary}
\end{center}}
\end{footnotesize}
\end{table*}

We also tested RGA (standard and its elite version) 
with a rigid bound consideration in its operators. In this case, the
probability distributions of both
{SBX} and polynomial mutation operator are changed in a way so as to
always create a feasible solution. 
It is found that the Standard-RGA with rigid bounds shows 
convergence only when optimum is on the boundary (Table~\ref{tab:rga-standard-rigid-boundary}). 
The performance of Elite-RGA with rigid bounds is slightly better 
(Table~\ref{tab:rga-elite-rigid-boundary}).
Overall, SBX operating within the rigid bounds is found to perform slightly better
compared to the RGAs employing boundary-handling mechanisms. 
However, as mentioned earlier, in the scenarios where the generation scheme cannot 
guarantee creation of feasible only solutions there is a necessary
need for constraint-handling strategies. 

\begin{table*}[htb]
\begin{footnotesize}
\begin{minipage}[b]{1.0\linewidth}
\caption{Elitist-RGA with rigid boundary with termination criterion $10^{-10}$.}
\begin{center}
\begin{tabular}{|lrrr|} \hline
{{Strategy}}     & {{Best}}& {{Median}}& {{Worst}} \\ \hline \hline
                \multicolumn{4}{|c|}{ Optimum on the boundary}                      \\      
$F_{elp}$               & 7,300 & 7,900 	& 8,400    	\\    
$F_{sch}$               & 6,500 & 6,900 & 7,500    	 \\    
$F_{ack}$               & 9,400 & 10,400 & 12,200    	\\     
$F_{ros}$               & 11000 (10) & 12700 & 16400      \\      \hline  \hline
\multicolumn{4}{|c|}{ Optimum in the center }               \\   
\multicolumn{4}{|c|}{ {\textit{1.24e-01(DNC)}}}               \\
\hline \hline   
\multicolumn{4}{|c|}{ Optimum close to the boundary edge}      \\    
$F_{elp}$               & 579,800 (3)  &885,900 & 908,600    \\      
$F_{sch}$               & \textit{2.73E-00} \textit{DNC} & \textit{6.18E-00} & \textit{1.34E-00}   \\  
$F_{ack}$               & \textit{1.75E-01}  \textit{DNC} & \textit{8.38E-01} & \textit{2.93E-00}  \\  
$F_{ros}$               & \textit{3.29E-00}  \textit{DNC} & \textit{4.91E+00} & \textit{5.44E+00}  \\      \hline 
\end{tabular}
\label{tab:rga-elite-rigid-boundary}
\end{center}
\end{minipage}
\end{footnotesize}
\end{table*}
\section{Higher-Dimensional Problems}\label{sec:scale-up}
\begin{figure}[hbt]
\begin{center}
\includegraphics[scale=1.0]{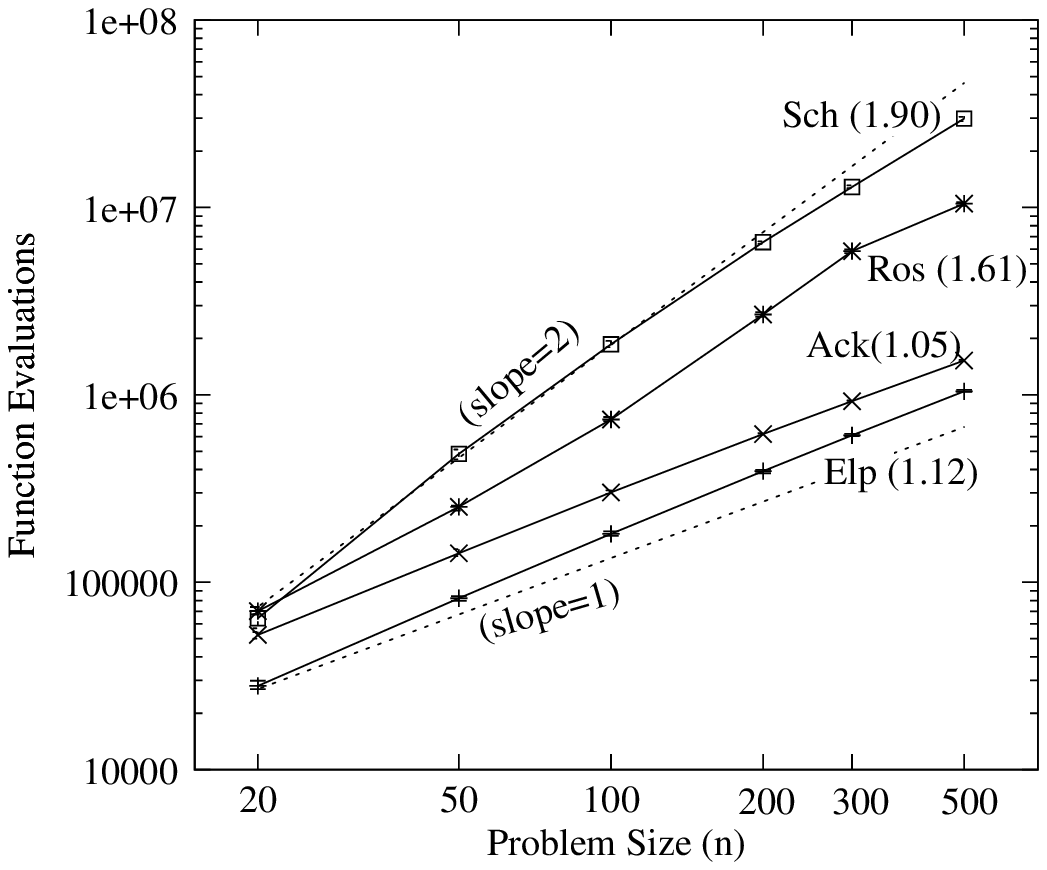}    
\end{center}
\caption{Scale-up study on four test problems for the DE algorithm with
  the proposed IP-S approach shows sub-quadratic performance in all
  problems. Slopes of a fitted polynomial line is marked within
  brackets for each problem. Linear and quadratic limit lines are shown with dashed lines.}
\label{fig:DE-scale-up}
\end{figure}

As the dimensionality of the search space increases it becomes
difficult for a search algorithm to locate the optimum.
Constraint-handling methods play even a more critical role
in such cases.  
So far in this study, DE has been found to be the best algorithm. Next, we consider all
four unimodal test problems with an increasing problem size: $n=20$, $50$, $100$, $200$, $300$,
and $500$. For all problems the variable bounds were chosen such that optima occured
near the center of the search space. No population scaling is used for DE. The DE parameters are chosen as $F=0.8$ and $CR=0.9$.
For $\alpha = 1.2$ we were able to achieve a high degree of convergence and
results are shown in Figure~\ref{fig:DE-scale-up}.
As seen from the figure, although it is expected that the required
number of function evaluations would increase with the 
number of variables, the increase is sub-quadratic.
Each case is run $20$ times and the termination criteria
is set as $10^{-10}$. All 20 runs are found to be successful in each case,
demonstrating the robustness of the method in terms of finding the
optimum with a high precision.
Particularly problems with large variables, complex search spaces and 
highly nonlinear constraints, such a methodology should be useful in terms of 
applying the method to different real-world problems.  

It is worth mentioning that authors independently tested other scenarios for 
large scale problems with corresponding optimum on the boundary, and in order to achieve convergence
with IP methods we had to significantly reduce the 
values of $\alpha$. Without lowering $\alpha$, particularly, PSO did not show any convergence. 
As expected in larger dimensions the probability to sample a point closer to the boundary
decreases and hence a steeper distribution is needed. However, this highlights the usefulness
of the parameter $\alpha$ to modify the behavior of the IPMs so as to yield the desired performance. 

\section{General Purpose Constraint-Handling} 
\label{sec:Constraint-Programming}

So far we have carried out simulations on problems where constraints have 
manifested as the variable bounds. The IP methods proposed in this paper can be easily
extended and employed for solving nonlinear constrained optimization problems (inclusive of 
variable bounds).\footnote{By \textit{General Purpose Constraint-Handling} we imply
tackling of all variable bounds, inequality constraints and equality constraints.}

As an example, let us consider a generic inequality constraint function: $g_j(\vec{x})
\geq 0$ -- the $j$-th constraint in a set of $J$ inequality
constraints. In an optimization algorithm, every created (offspring) solution
$\vec{x}^c$ at an iteration must be
checked for its feasibility. If
$\vec{x}^c$ satisfies all $J$ inequality constraints, the solution is
feasible and the algorithm can proceed with the created solution. 
But if $\vec{x}^c$ does not
satisfy one or more of $J$ constraints, the solution is
infeasible and should be repaired before proceeding further. 

\begin{figure}[hbt]
\begin{center}
\includegraphics[scale=1.0]{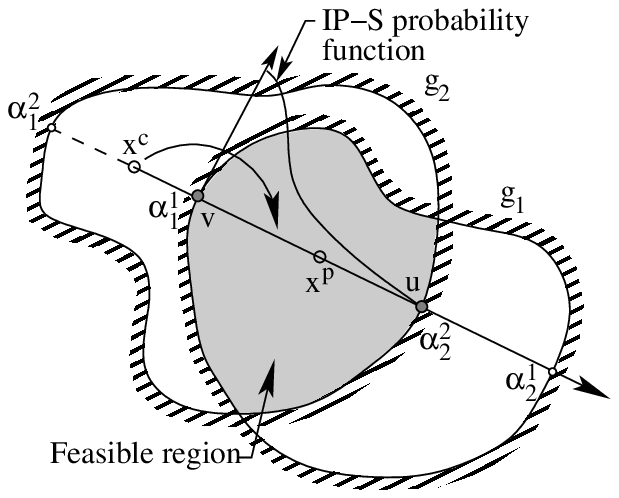}    
\caption{Extension of constraint-handling approaches to constrained optimization problems.}
\label{fig:constr}
\end{center}
\end{figure}

Let us
illustrate the procedure using the inverse parabolic (IP) approach described in
Section~\ref{sec:IP}; though other constraint-handling
methods discussed before may also be used. The IP approaches require us to locate
intersection points $\vec{v}$ and $\vec{u}$: two bounds in the
direction of ($\vec{x}^p-\vec{x}^c$), where $\vec{x}^p$ is one of the
parent solutions that created the offspring solution (see Figure~\ref{fig:constr}). The critical intersection
point can be found by
finding multiple roots of the direction vector with each constraint
$g_j(\vec{x})$ and then choosing the
smallest root. 
We define a parameter
$\alpha$ as the extent of a point from $\vec{x}^c$, as
follows:
\begin{equation}
\vec{x}(\alpha) = \vec{x}^c + \alpha
\frac{\vec{x}^p-\vec{x}^c}{\|\vec{x}^p-\vec{x}^c\|}.
\label{eq:mapping}
\end{equation}

Substituting above expression for $\vec{x}(\alpha)$ \footnote{Note that $\alpha$
here for calculating points should not be confused with parameter $\alpha$ introduced in
the proposed constraint-handling methods.}in the $j$-th constraint function, we have the following
root-finding problem for the $j$-th constraint:
\begin{equation}
g_j(\vec{x}(\alpha)) = 0.
\end{equation}
Let us say the roots of the above equation are $\alpha_k^j$ for
$k=1,2\ldots,K_j$. The above procedure can now be repeated for all $J$
inequality constraints and corresponding roots can be found one at a time. 
Since the extent of $\alpha$ to reach $\vec{x}^p$ from $\vec{x}^c$
is given as 
\[\alpha^p = \|\vec{x}^p-\vec{x}^c\|,\]
we are now ready to compute the two closest bounds (lower and upper bounds) on
$\alpha$ for our consideration, as
follows:
\begin{eqnarray}
\alpha^v &=& \max \{\alpha_k^j | 0 \leq \alpha_k^j \leq \alpha^p,
 \forall
k, \forall j\}, \\ 
\alpha^u &=& \min \{\alpha_k^j | \alpha_k^j \geq \alpha^p, \forall
k, \forall j\}.
\end{eqnarray}
IP-S and IP-C approaches presented in Section~\ref{sec:IP} now can be used to map the violated variable
value $x_i^c$ into the feasible region using $d=\alpha^v$ (the lower bound),
$d_u=\alpha^u$ (the upper bound) and $d^p=\alpha^p$ (location of parent). 
It is clear that the only difficult aspect of this method is to find
multiple intersection points in presence of nonlinear constraints. 

For the sake of completeness, we show here that the two bounds for each variable: $x_i\geq x_i^{(L)}$ and
$x_i\leq x_i^{(U)}$ used in previous sections can be also be treated
uniformly using the above described approach. The two bounds can be written together as follows:
\begin{equation}
\left(x_i-x_i^{(L)}\right)\left(x_i^{(U)}-x_i\right) \geq 0.
\end{equation}
Note that a simultaneous non-positive value of each of the bracketed terms
is not possible, thus only way to satisfy the above left side is to
make each bracketed term non-negative. The above inequality can be
considered as a quadratic constraint function, instead of two
independent variable bounds and treated as a single combined nonlinear
constraint and by finding both roots of the resulting quadratic root-finding equation.  

Finally, the above procedure can also be extended to handle equality constraints ($h_k(\vec{x})=0$) with a
relaxation as follows: $-\epsilon_k \leq h_k(\vec{x}) \leq
\epsilon_k$. Again, they can be combined together as follows:
\begin{equation}
\epsilon_k^2-(h_k(\vec{x}))^2 \geq 0.
\end{equation}
Alternatively, the above can also be written as $\epsilon_k -
|h_k(\vec{x})| \geq 0$ and may be useful for non-gradient based
optimization methods, such as evolutionary algorithms. We now show the
working of the above constraint handling procedure on a number of
constrained optimization problems. 
 
\subsection{Illustrations on Nonlinear Constrained Optimization}

First, we consider the three unconstrained problems used in previous
sections as $f(\vec{x})$, but now add an inequality constraint by imposing a
quadratic constraint that makes the solutions fall within a radius of one-unit from
a chosen point $\vec{o}$:
\begin{equation}
\begin{array}{rl}
\mbox{Minimize}  & f(\vec{x}), \\
\mbox{subject to} & \sum_{i=1}^{n} (x_i-o_i)^2 \leq 1.	
\end{array}
\label{eq:convex-opti-problem}
\end{equation}
There are no explicit variable bounds in the above problem. 
By choosing different locations of the center
of the hyper-sphere ($\vec{o}$), we can have different scenarios of
the resulting constrained optimum. 
If the minimum of the objective function (without constraints)
lies at the origin, then setting $o_i=0$ the unconstrained minimum is also the solution to the
constrained problem, and this case is similar to the ``Optimum at the
Center'' (but in the context
of constrained optimization now).
The optimal solution is at $x_i=0$ with $f^*=0$. DE with IP-S and previous parameter 
settings is applied to this new constrained problem, and the results from 50 different runs for this case are shown in
Table~\ref{tab:convex-constraints-de-1}.
\begin{table*}[ht]
\begin{footnotesize}
\begin{center}
\caption{Results for test functions with DE for $o_i=0$ with
  $S=10^{-10}$.}
\label{tab:convex-constraints-de-1}   
\begin{tabular}{|lrrr|} \hline
{{Strategy}} & {{Best}} & {{Median}} & {{Worst}}     \\\hline \hline
$F_{\rm elp}$             & 22,800 &	23,750 &	24,950 		  \\ 
$F_{\rm sch}$          &      183,750	& 206,000 &	229,150                          \\
$F_{\rm ack}$            & 42,800 &	44,250	& 45,500 	                      \\ \hline
\end{tabular}
\end{center}
\end{footnotesize}
\end{table*}

As a next case, we consider $o_i = 2$. The constrained
minimum is now different from that of the unconstrained problem, as
the original unconstrained minimum is no more feasible. This case is equivalent to ``Optima on 
the Constraint Boundary''. Since the optimum value is not zero as before, instead of choosing
a termination criterion based on $S$ value, we allocate a maximum of
one million function evaluations for a run
and record the obtained optimized solution. The {best fitness} values for 
$f(\vec{x})$ as $f_{elp}$, $f_{sch}$ and $f_{ack}$ are shown in Table
~\ref{tab:convex-constraints-de-2}. For each function, we verified
that the obtained optimized solution satisfies the KKT optimality
conditions \cite{rekl,debOptiBOOK} suggesting that a truly optimum solution has
been found by this procedure. 
\begin{table*}[ht]
\begin{footnotesize}
\begin{center}
\caption{Best fitness results for test functions with DE for $o_i=2$.}
\begin{tabular}{|ccc|} \hline
$F_{\rm elp}$             & $F_{\rm sch}$& $F_{\rm ack}$	 		  \\ \hline \hline
      640.93 $\pm$ 0.00  & 8871.06 $\pm$ 0.39	& 6.56 $\pm$ 0.00 	                         \\ \hline
\end{tabular}
\label{tab:convex-constraints-de-2}      
\end{center}
\end{footnotesize}
\end{table*}
  
Next, we consider two additional nonlinear constrained optimization
problems (TP5 and TP8) from \cite{debpenalty} and a well-studied structural design and mechanics problem (`Weld') taken from \cite{ragsdell1976optimal}.
The details on the mechanics of the welded structure and the beam deformation can be found in \cite{shigley1963engineering,timoshenko1962theory}.
These problems have 
multiple nonlinear inequality constraints and our goal is to demonstrate the performance of our proposed constraint-handling
methods. We used DE with
IP-S method with the following 
parameter settings: $NP=50$, $F=0.7$, $CR=0.5$, and $\alpha=1.2$ for all three problems.
A maximum of 200,000 function evaluations were allowed and a termination criteria of $10^{-3}$
from the known optima is chosen. 
The problem definitions of TP5, TP8 and `Weld' are as follows:\\

\noindent \textbf{TP5:}
\begin{equation}
\label{eq:TP5}
\begin{array}{rl}
\mbox{Minimize} & f(\vec{x}) = (x_1-10)^2+5(x_2-12)^2+x_3^4+3(x_4-11)^2  \\	
& \qquad	+ 10x_5^6+7x_6^2+x^4_7-4x_6x_7-10x_6-8x_7, \\
\mbox{subject to} & g_1(\vec{x})  \equiv 2x_1^2+3x_2^4+x_3+4x_4^2+5x_5 \leq 127,	\\
& g_2(\vec{x}) \equiv   7x_1+3x_2+10x_3^2+x_4-x_5 \leq 282, \\	
& g_3(\vec{x})  \equiv    23x_1+x_2^2+6x_6^2-8x_7 \leq 196, \\
& g_4(\vec{x}) \equiv 4x_1^2+x_2^2-3x_1x_2+2x_3^2 +5x_6-11x_7 \leq 0,\\		
& -10 \leq x_i \leq 10,   \quad  i = 1,\ldots,7. 
\end{array} 
\end{equation}  

\noindent\textbf{TP8:}
\begin{equation}
\label{eq:TP8}
\begin{array}{rl}
\mbox{Minimize} & f(\vec{x}) = x_1^2+x_2^2+x_1x_2-14x_1-16x_2+2(x_9-10)^2\\
 & \qquad + 2(x_6-1)^2 + 5x_7^2+7(x_8-11)^2+45+(x_{10}-7)^2 \\
 & \qquad + (x_3-10)^2 + 4(x_4-5)^2 + (x_5-3)^2\\
\mbox{subject to} & g_1(\vec{x})  \equiv 4x_1+5x_2-3x_7+9x_8 \leq 105,\\
& g_2(\vec{x})  \equiv	10x_1-8x_2-17x_7+2x_8 \leq 0,\\
& g_3(\vec{x})  \equiv -8x_1+2x_2+5x_9-2x_{10}  \leq 12,\\
& g_4(\vec{x})\equiv  3(x_1-2)^2 + 4(x_2-3)^2+2x_3^2-7x_4 \leq 120,\\
& g_5(\vec{x})	\equiv 5x_1^2+8x_2+(x_3-6)^2-2x_4 \leq 40,	\\
& g_6(\vec{x})	\equiv	 x_1^2+2(x_2-2)^2-2x_1x_2+14x_5-6x_6 \leq 0,\\
& g_7(\vec{x})	 \equiv	 0.5(x_1-8)^2+2(x_2-4)^2+3x_5^2-x_6  \leq 30,\\
& g_8(\vec{x})	 \equiv	 -3x_1+6x_2+12(x_9-8)^2-7x_{10} \leq 0,\\
& -10 \leq x_i\leq 10, \quad i = 1,\ldots,10. 
\end{array} 
\end{equation}	 

\noindent\textbf{`Weld':}
\begin{equation}
\label{eq:weld-problem}
\begin{array}{rl}
\mbox{Minimize} & f(\vec{x}) = 1.10471h^2l+0.04811tb(14.0+l),	 \\	
\mbox{subject to} & g_1(\vec{x}) \equiv  \tau(\vec{x}) \leq 13,600,   \\
& g_2(\vec{x})  \equiv  \sigma(\vec{x}) \leq 30,000,	\\
& g_3(\vec{x})  \equiv  h-b \leq 0, \\	 
& g_4(\vec{x})  \equiv  P_c(\vec{x})  \geq 6,000,\\
& g_5(\vec{x})  \equiv  \delta(\vec{x}) \leq 0.25,		\\
& 0.125\leq (h,b) \leq 5, \mbox{ and } 0.1 \leq (l,t) \leq 10, 
\end{array}
\end{equation}
where,			 
\begin{eqnarray*}
\tau(\vec{x})	&=&  \sqrt{ (\tau')^2 + (\tau'')^2 + (l\tau'\tau'')/ \sqrt{0.25(l^2+(h+t)^2)}}, \\		\\	
\tau' &=& \frac{6,000}{\sqrt{2}hl}, \\
\tau'' &=& \frac{6,000(14+0.5l)\sqrt{0.25*(l^2+(h+t)^2)}}{2[0.707hl(l^2/12+0.25(h+t)^2)] }, \\
\sigma(\vec{x}) &=& \frac{504,000}{t^2b}, \\
\delta(\vec{x}) &=& \frac{2.1952}{t^3b}, \\
P_c(\vec{x}) &=& 64,746.022(1-0.0282346t)tb^3.   
\end{eqnarray*} 
 
The feasible region in the above problems is quite complex, unlike hypercubes in case
of problems with variable bounds, and since our methods require feasible initial population,
we first identified a single feasible-seed solution (known as the seed solution), and generated other several other 
random solutions. Amongst the several randomly generated solutions, those infeasible,
were brought into the feasible region using IP-S method and feasible-seed solution as the reference.
The optimum results for TP5, TP8 and `Weld', thus found, are shown in the
Table~\ref{tab:Non-linear-opti}. 

\begin{table*}[htb]
\caption{Results from TP5, TP8 and `Weld' problem. For each problem, the
  obtained solution also satisfies the KKT conditions.}
\label{tab:Non-linear-opti}
\begin{center}
\begin{footnotesize}
\begin{tabular}{|l|l|l|l|} \hline
	 	&  \multirow{2}{1.5cm}{Optimum Value ($f^*$)}& {Corresponding Solution Vector ($\vec{x}^{\ast}$)}
                & \multirow{2}{1.2cm}{Active Constraints} \\ 
 & & &  \\ \hline
\textbf{TP5}	&    $680.63$
&$(2.330,1.953,-0.473,4.362,-0.628,1.035,1.591)^T$	&  $g_1$, $g_4$   				\\      \hline
\textbf{TP8}		& $24.33$ &
$(2.160,2.393,8.777,5.088,0.999,1.437,1.298,9.810,8.209,8.277)^T$&
$g_1$ to $g_6$\\      \hline  
\textbf{`Weld'}	&
$2.38$& $(0.244,6.219,8.291,0.244)^T$	& $g_1$ to $g_4$  \\      \hline  
\end{tabular}
\end{footnotesize}
\end{center}
\end{table*}

To verify the obtained optimality of our solutions, we employed MATLAB sequential
quadratic programming (SQP) toolbox with a
termination criterion of $10^{-6}$ to solve each of the three problems. The solution obtained from SQP
method matches with our reported solution indicating that our proposed constrained
handling procedure is successful
in solving the above problems.

\begin{figure}[hbt]
\begin{center}
\includegraphics[scale=.35,angle=-90]{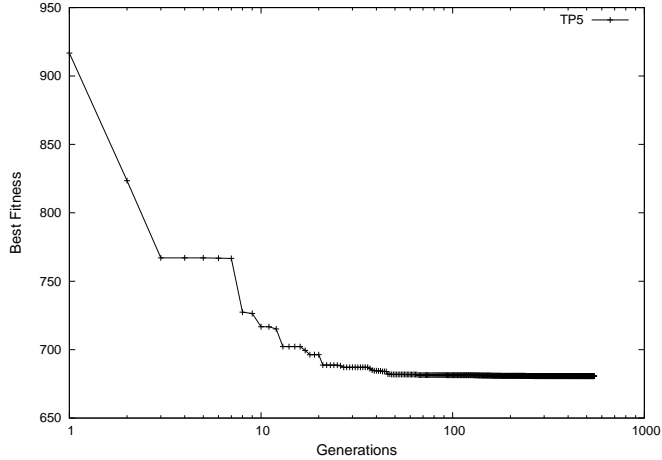}    
\end{center}
\caption{Convergence plot for TP5.}
\label{fig:TP5}
\end{figure}

\begin{figure}[hbt]
\begin{center}
\includegraphics[scale=.35,angle=-90]{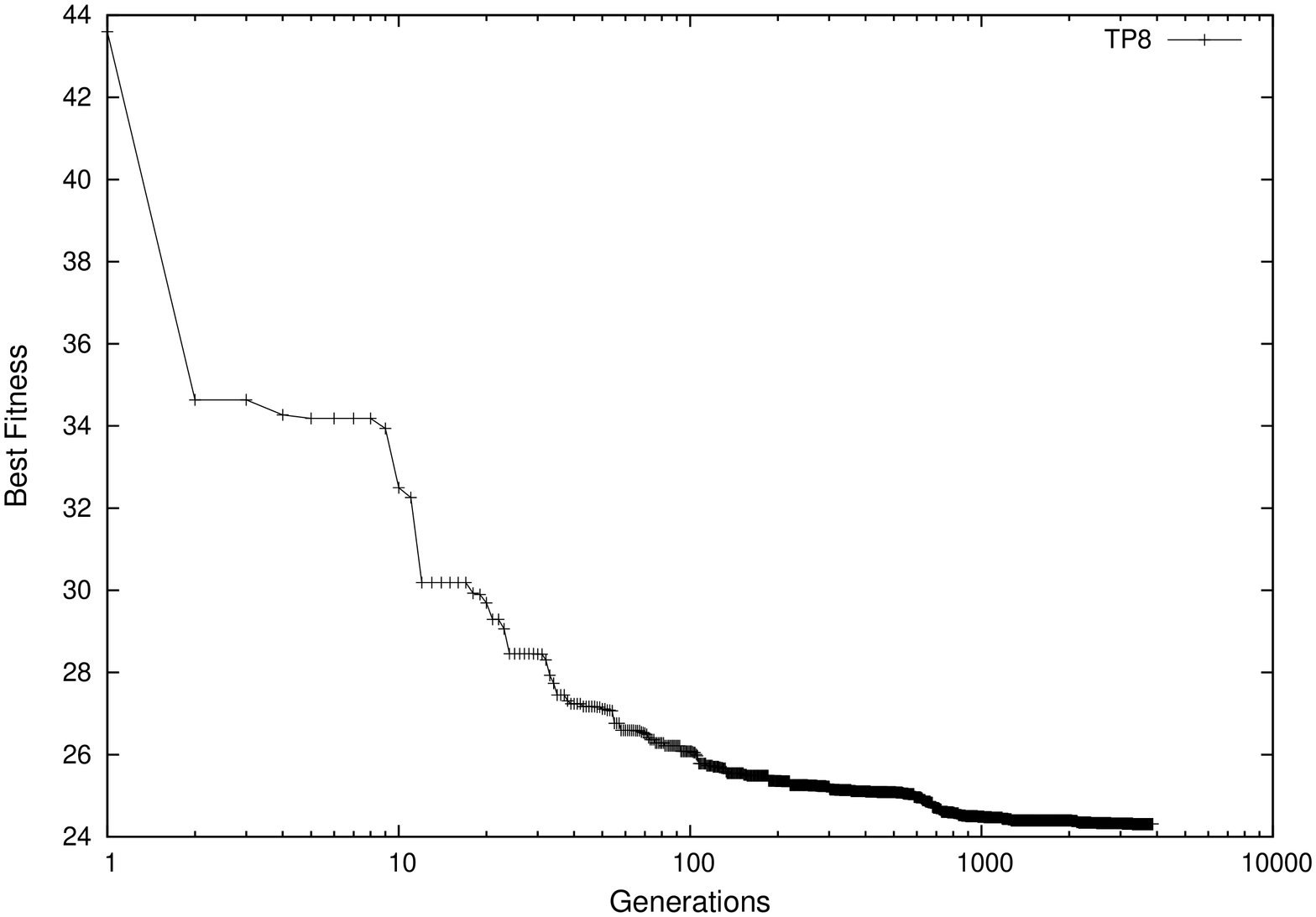}    
\end{center}
\caption{Convergence plot for TP8.}
\label{fig:TP8}
\end{figure}

\begin{figure}[hbt]
\begin{center}
\includegraphics[scale=.35,angle=-90]{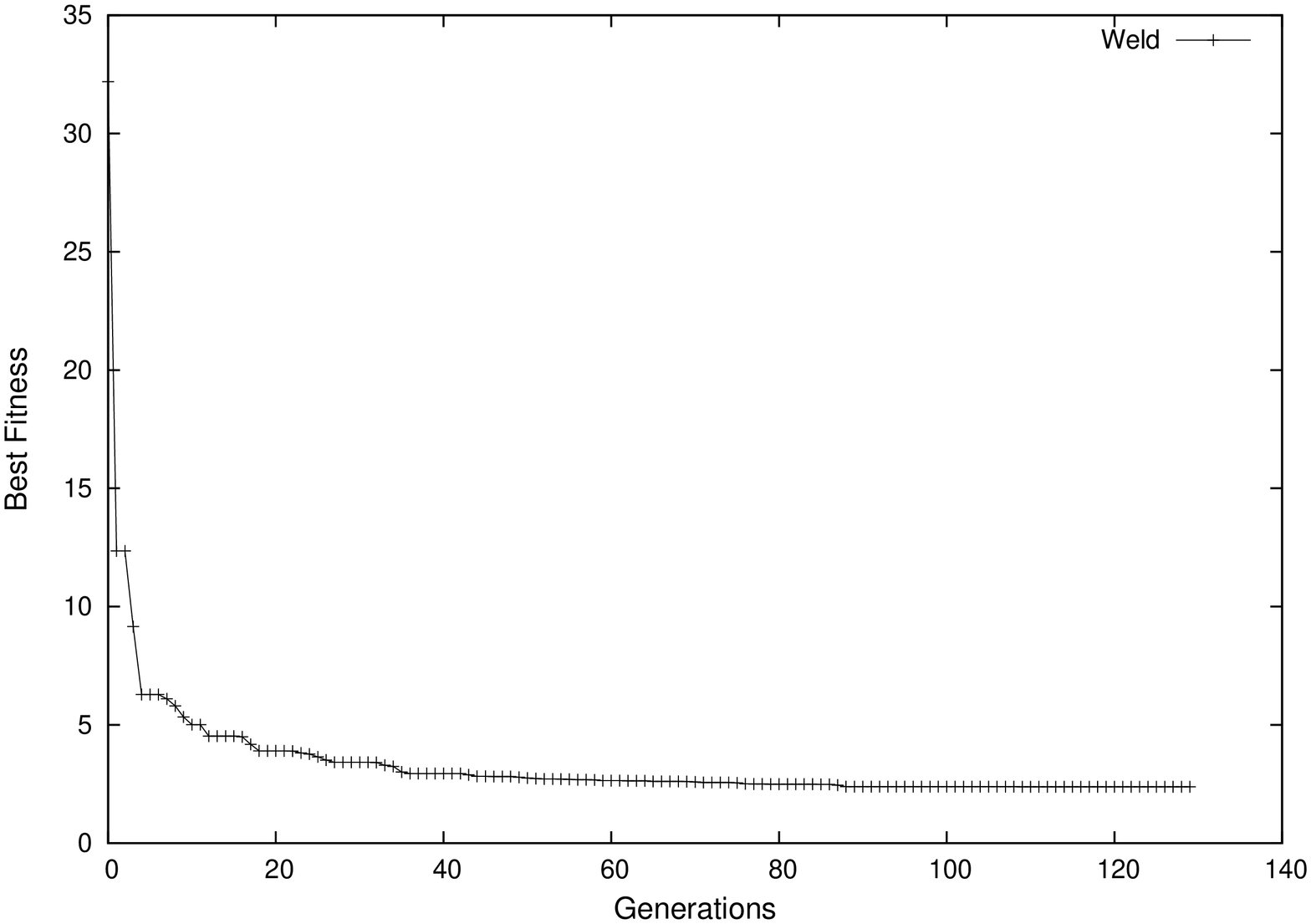}    
\end{center}
\caption{Convergence plot for TP8.}
\label{fig:WELD}
\end{figure}

Finally, the convergence plots of a sample run on these test
problems is shown in Figures~\ref{fig:TP5}, \ref{fig:TP8}, and \ref{fig:WELD}, respectively.
From the graphs it is clear that our proposed method is able to
converge to the final solution quite effectively.

\section{Conclusions}
\label{sec:Conclusion}
The existing constraint-handling strategies that repair solutions by bringing
them back into the search spaces exhibit several
inadequacies. This paper has addressed the task of studying, and proposing two new, explicit feasibility preserving
constraint-handling methods for real-parameter optimization using evolutionary algorithms. 
Our proposed single parameter Inverse Parabolic Methods with stochastic and adaptive components, 
overcome limitations of existing methods and perform effectively
on several test problems.
Empirical comparisons on four different
benchmark test problems ($F_{elp}$, $F_{sch}$, $F_{ack}$, and $F_{ros}$) 
with three different settings of the optimum relative to the
variable boundaries revealed key insights into the performance of PSO, GAs and DE
in conjunction with different constraint-handling strategies.
It was noted that in addition to the critical role of constraint-handling strategy (which
is primarily responsible for bringing the infeasible solutions back into the search space),
the generation scheme (child-creation step) in an evolutionary algorithm must create efficient solutions 
in order to proceed effectively.
Exponential and Inverse Parabolic Methods were most robust methods and never
failed to solve any problem. The other constraint-handling strategies
were either too deterministic and/or operated without utilizing sufficient useful information from 
the solutions. The probabilistic methods
were able to bring the solutions back into the
feasible part of the search space and showed a consistent
performance. In particular, scale-up studies on four problems, up to 500 variables, demonstrated
sub-quadratic empirical run-time complexity with the proposed IP-S method.

Finally, the success of the proposed IP-S scheme is demonstrated 
on generalized nonlinear constrained optimization problems.   
For such problems, the IP-S method requires finding the lower and upper bounds
for feasible region along the direction of search by solving a series
of root-finding problems. 
To best of our knowledge, the
proposed constraint-handling approach is a first explicit feasibility preserving method that has demonstrated
success on optimization problems with variable bounds and nonlinear constraints.
The approach is arguably general, and can be 
applied with complex real-parameter search spaces such as non-convex, discontinuous, etc., 
in addition to problems dealing with multiple conflicting objectives, multi-modalities, dynamic
objective functions, and other complexities.
We expect that proposed constraint-handling scheme will find its utility in 
solving complex real-world constrained optimization problems using evolutionary algorithms. 
An interesting direction for the future would be to carry out one-to-one comparison between
evolutionary methods employing constrained handling strategies and the classical
constrained optimization algorithms. Such studies shall benefit the practitioners in optimization 
to compare and evaluate different algorithms in a unified frame-work.
In particular, further development of a robust, parameter-less
and explicit feasibility preserving constraint-handling procedure can be attempted.
Other probability distribution functions and utilizing of information 
from other solutions of the population can also be attempted. 
\section*{Acknowledgments}
Nikhil Padhye acknowledges past discussions with Dr. C.K. Mohan on swarm intelligence. 
Professor Kalyanmoy Deb acknowledges 
the support provided by the J. C. Bose National
fellowship generously provided by the Department of Science and
Technology (DST), Government of India. 
{\footnotesize
\bibliographystyle{plain}
\bibliography{references}
}
\end{document}